\pdfoutput=1

\documentclass{article}

\usepackage{PRIMEarxiv}

\usepackage[utf8]{inputenc}
\usepackage[T1]{fontenc}
\usepackage{hyperref}
\usepackage{url}
\usepackage[numbers,sort&compress]{natbib}
\usepackage{booktabs}
\usepackage{amsmath}
\usepackage{amssymb}
\usepackage{microtype}
\usepackage{graphicx}
\usepackage{array}
\usepackage{tabularx}
\usepackage{enumitem}
\usepackage{xcolor}
\usepackage{listings}
\usepackage{placeins}
\usepackage{caption}
\usepackage{ragged2e}
\renewcommand\cite{\citep}
\definecolor{primarycolor}{RGB}{36,87,122}
\AtBeginDocument{\fussy\raggedbottom\setlength{\emergencystretch}{3em}\normalsize}
\microtypesetup{expansion=false}

\newcommand{\code}[1]{{\ttfamily\def\_{\char95\relax\penalty200\relax}#1}}
\newcommand{\agenttune}{\textsf{AgentTune}}
\newcommand{\eventlog}{\code{EventLog}}

\hypersetup{
  pdftitle={AgentTune: A Unified Training Lifecycle for Tool-Using LLM Agents},
  pdfauthor={R E Zera Marveen Lyngkhoi and Vidushee Vats and Abhivansh Gupta and Vinay Kumar Sankarapu and Pratinav Seth},
  pdfsubject={Agentic LLM training, tool use, reinforcement learning, distillation},
  pdfkeywords={agentic LLMs, reinforcement learning, tool use, knowledge distillation, RAG, self-healing systems},
  colorlinks=false,
}

\definecolor{codebg}{RGB}{247,247,249}
\definecolor{codekw}{RGB}{0,90,140}
\definecolor{codestr}{RGB}{140,60,20}
\definecolor{codecom}{RGB}{110,110,110}
\lstdefinestyle{at}{
  language=Python,
  backgroundcolor=\color{codebg},
  basicstyle=\ttfamily\fontsize{8}{9.5}\selectfont,
  keywordstyle=\color{codekw},
  stringstyle=\color{codestr},
  commentstyle=\color{codecom}\itshape,
  showstringspaces=false,
  breaklines=true,
  breakatwhitespace=true,
  columns=fullflexible,
  keepspaces=true,
  aboveskip=6pt,
  belowskip=4pt,
}
\title{%
  \begin{minipage}[c]{0.30\textwidth}
    \centering
    \includegraphics[width=\linewidth]{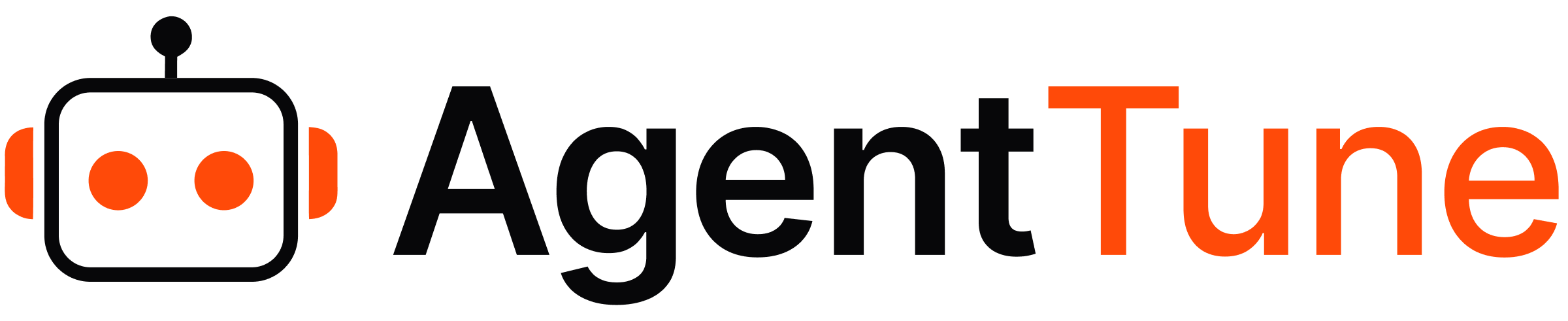}
  \end{minipage}%
  \hspace{1em}%
  \begin{minipage}[c]{0.60\textwidth}
    \raggedright
    A Unified Training Lifecycle\\
    for Tool-Using LLM Agents
  \end{minipage}%
}

\author{
  R E Zera Marveen Lyngkhoi\thanks{These authors contributed equally.} \quad
  Abhivansh Gupta\footnotemark[1] \quad
  Vidushee Vats\footnotemark[1] \\  Ram Mohan Rao Kadiyala \\
  Vinay Kumar Sankarapu \quad Pratinav Seth \\
  \affiliation{Lexsi Labs}
}
\setabstract{%
\justifying
Tool-using agents increasingly require a common substrate for building, training, evaluating, distilling, and maintaining agentic systems, yet existing tooling typically addresses these stages in isolation. We present \agenttune{}, a source-available Python library that provides a unified framework for developing and optimizing tool-using agents. At its core, AgentTune introduces a two-tier \eventlog{} representation that preserves agent trajectories while supporting consistent projections for training, evaluation, and auditing. Building on this substrate, the library provides a shared agent runtime, tool-calling reinforcement learning, agentic distillation, retrieval-augmented generation, decision workflows, evaluation utilities, and a designed path toward self-healing agents. These components are exposed through composable interfaces so that agent strategies, training methods, evaluation procedures, and recovery mechanisms can be combined without requiring separate representations or pipelines. We demonstrate the framework across agentic tasks including legal contract search and domain-specific applications, illustrating its use for trajectory-based optimization, tool-use evaluation, and cross-task agent development. AgentTune is released as an open-source library at \url{https://github.com/Lexsi-Labs/AgentTune}.
}

\setkeywords{Agentic LLMs, Reinforcement Learning, Tool Use, Knowledge Distillation, Retrieval-Augmented Generation, Self-Healing Systems, Source-Available Libraries}

\runningtitle{\agenttune{}: A Unified Training Lifecycle for Tool-Using LLM Agents}

\begin{document}
\maketitle

\section{Introduction}
\label{sec:introduction}

An LLM agent repeatedly reasons, calls a tool, reads the environment response, and decides whether to continue. Execution frameworks make that loop straightforward, but a useful agent system must also preserve the decisions and tool interactions needed for evaluation, training, compression, and recovery. Existing tools specialize in individual stages: TRL \citep{vonwerra2020trl}, verl \citep{sheng2024hybridflow} and OpenRLHF \citep{hu2024openrlhf} train policies, LangGraph \citep{langgraph2024} orchestrates stateful agent loops, the \code{lm-eval} harness \citep{eval-harness} standardizes static evaluation, and a separate literature covers distillation \citep{hinton2015distilling,gu2024minillm}. The fragmentation is at their interfaces: an RL rollout, a decision-workflow state, and an evaluation trace use different representations, so scoring, loss masking, failure detection, and audit logic are repeatedly reconstructed.

We distinguish \agenttune{} from AgentTuning \citep{Zeng2023AgentTuningEG}: AgentTuning is a method and dataset for fine-tuning LLMs on agent interaction trajectories, whereas \agenttune{} is a software library for collecting, representing, training, evaluating, distilling, and recovering tool-using agents. The common trajectory substrate is therefore complementary rather than a claim of methodological equivalence.

\agenttune{} is organized around one interface between these stages: the two-tier \eventlog{}. Agent trajectories from the runtime, RL rollouts, DECIDE workflows, and evaluation traces project into the schema; the same record projects out to SFT rows, evaluator inputs, and audit records. The library is designed to support the agent development lifecycle from build and collection through evaluation, training, distillation, and recovery, but the paper evaluates these capabilities separately rather than claiming a single experiment demonstrates the entire lifecycle. The legal case study provides the strongest task-level evidence: a search agent is trained on synthesized questions and then evaluated on three unseen corpora (\S\ref{sec:legal}).

\paragraph{Contributions.} This paper makes four contributions:
\begin{itemize}[leftmargin=*,itemsep=1pt,topsep=3pt]
\item \textbf{A unified trajectory contract.} The two-tier \eventlog{} represents agent interactions once and provides projections for training, evaluation, and audit, including segment-aware completion-only loss masking (\S\ref{sec:eventlog}).
\item \textbf{A shared agent-training stack.} The agentic spine and TRL-backed RL core provide a common runtime and training interface for tool-using agents across GRPO, PPO, DPO, RLOO, and BCO, with agentic distillation and retrieval training using the same trajectory substrate (\S\ref{sec:spine},~\S\ref{sec:rlcore}).
\item \textbf{Decision workflows and recovery.} DECIDE provides YAML-defined workflows and audit trails, while the architecture supports converting detected failures into corrective training data and gated retraining/deployment (\S\ref{sec:decide},~\S\ref{sec:selfheal}).
\item \textbf{Empirical evidence across representative paths.} GPU-verified smoke tests establish that the principal training paths execute end to end, while the legal contract-search study provides the strongest task-level result: GRPO improves held-out CUAD exact match from 60.0\% to 69.0\% and reduces searches per question from 3.00 to 2.90 (\S\ref{sec:casestudies},~\S\ref{sec:legal}).
\end{itemize}

\section{Installation and Getting Started}
\label{sec:install}

\paragraph{Release and license.}
\agenttune{} is released under the Lexsi Labs Source Available License (LSAL) v1.1: free for
research, evaluation, education and audit, with commercial use and safety-behavior removal restricted.

\paragraph{Requirements.}
The trainer factory derives its accepted parameter surface from the installed TRL by
\code{inspect.signature()}, so a different TRL release accepts a different set of kwargs. The
repository records the exact dependency set the captured runs used.

One command installs the agentic spine, the RL core, the RAG package, the extra evaluation metrics, the PostgreSQL DECIDE destination, and the OpenEnv sandbox:

\begin{lstlisting}[language=bash,caption={Install from source. \code{[service]}, \code{[docs]}, \code{[dev]}, and \code{[flash-attn]} are optional extras.}]
git clone https://github.com/Lexsi-Labs/AgentTune.git
cd AgentTune
pip install -e .
\end{lstlisting}

The spine core (\code{agenttune.agentic}) needs no GPU, no model download, no network and no API key; it does import PyTorch at package-import time, so the install is not lightweight even when nothing is trained. Training a model with any RL algorithm needs PyTorch and a CUDA GPU. Listing~\ref{lst:quickstart} builds one agent, runs it once, and evaluates it, with a scripted policy and an in-process tool, so it runs with no GPU, no network, and no model download.

\begin{lstlisting}[caption={The GPU-free quickstart: one agent, one episode, evaluation, all with a scripted policy and an in-process tool. The first two calls need no optional dependency; \code{evaluate\_agentic} additionally imports the trajectory evaluator, which pulls in the LLM-gateway client.}, label={lst:quickstart}]
from agenttune.agentic import Project, DictToolHarness, ReActStrategy

harness  = DictToolHarness({"add": lambda a, b: a + b}, max_steps=4)
strategy = ReActStrategy(
    lambda s: {"name": "add", "arguments": {"a": 2, "b": 3}}
    if s.step == 0 else {"name": "finish", "arguments": {"answer": "5"}})

proj = Project(strategy=strategy, harness=harness)
proj.infer("what is 2+3?")                                 # one episode -> EventLog
proj.evaluate([{"task": "what is 2+3?", "expected": "5"}]) # -> mean_score 1.0
proj.evaluate_agentic(["what is 2+3?"])                    # programmatic metrics
\end{lstlisting}

Every call in Listing~\ref{lst:quickstart} produces or consumes the same \eventlog{}. Replacing the scripted policy with a real generation function leaves the later training, distillation, and recovery stages unchanged. Section~\ref{sec:rlcore} shows a real GRPO run.

\section{Library Structure}
\label{sec:structure}

This section describes the library from the outside in. \S\ref{sec:scope} fixes what \agenttune{} covers and what it does not; \S\ref{sec:hla} maps the source tree and places the pieces around the \eventlog{}; and \S\ref{sec:eventlog} onwards takes each capability in turn, starting with the schema the rest depend on. Appendix~\ref{sec:examples} collects eight worked examples that compose the pieces described here.

\subsection{Scope and definitions}
\label{sec:scope}

Before describing the architecture, we fix the scope and the terms. \agenttune{} is organized as the trainer and the spine layered on it, plus DECIDE as an independent third surface, all sharing one command-line interface (Figure~\ref{fig:arch}). The map covers the paths that produce or consume trajectories; the scripted domain scenarios (\code{agenttune.scenarios}) and the dataset and diagnostic helpers (\code{agenttune.data}, \code{agenttune.utils}) are packaging around those paths and are out of scope here. The agentic spine, implemented in \code{agenttune.agentic}, combines an \code{AgentStrategy} with a \code{Harness} and runs the pair through a \code{Project}; every lifecycle method reads or writes an \eventlog{}. The shared RL core exposes \code{create\_agentic\_trainer}, a standalone factory for TRL-backed GRPO, PPO, DPO, RLOO, and BCO trainers with tool-calling rollouts. It is called directly by the public training wrapper, the RAG package, multi-agent orchestration, and DECIDE's retraining path.

DECIDE, implemented in \code{agenttune.decide}, compiles YAML decision workflows, appends each run to \code{audit.jsonl}, and uses that audit trail for closed-loop recovery. The \code{agenttune} command exposes training, pipeline, version, and DECIDE operations without requiring Python integration; it does not yet expose tool-calling training, because there is no \code{-{}-tools} flag, so agentic training is reachable only through the Python API (\S\ref{sec:limitations}). Appendix~\ref{app:glossary} collects the terms used throughout.

The spine and DECIDE share the RL core and the \eventlog{}/audit boundary. The remaining coupling is one-directional and deliberately thin: DECIDE never imports the spine, and the spine imports only DECIDE's standard-library dataclass contracts, so neither subsystem drags the other's execution machinery into a process that does not need it.

\paragraph{Where each capability lives.}
Table~\ref{tab:packages} maps the eight subsections that follow onto the source tree; they run in order from \S\ref{sec:eventlog}, schema first, because the rest depend on it. These layers correspond to the four contributions in \S\ref{sec:introduction}; the remaining utilities and case studies provide supporting implementation and empirical evidence. The layers are additive rather than a fixed stack: each speaks \eventlog{}, while evaluation cuts across them and is reachable from both the Python API and CLI (\S\ref{sec:evalimpl},~\S\ref{sec:packaging}).

\paragraph{Non-goals.}
\agenttune{} does not claim: (1) new optimizers --- it dispatches to TRL's trainers and inherits their single-node scaling limits, leaving cluster-scale RLHF to verl and OpenRLHF (\S\ref{sec:related},~\S\ref{sec:limitations}); (2) a new retrieval algorithm --- the RAG package supplies the trajectory substrate a search policy is trained on, not a ranking method (\S\ref{sec:related}); (3) a security sandbox --- the shipped tool defaults favor usability over confinement, and untrusted code belongs behind a remote sandbox (\S\ref{sec:ethics}); (4) certified decision systems --- the bundled domain templates demonstrate the workflow engine and encode plausible rather than regulator-approved logic (\S\ref{sec:ethics}); or (5) benchmark leadership --- no number in this paper is scored by an external agentic harness (\S\ref{sec:limitations}).

\subsection{Package layout and high-level architecture}
\label{sec:hla}

\begin{figure}[tb]
\centering
\IfFileExists{figures/architecture.png}{\includegraphics[width=\textwidth]{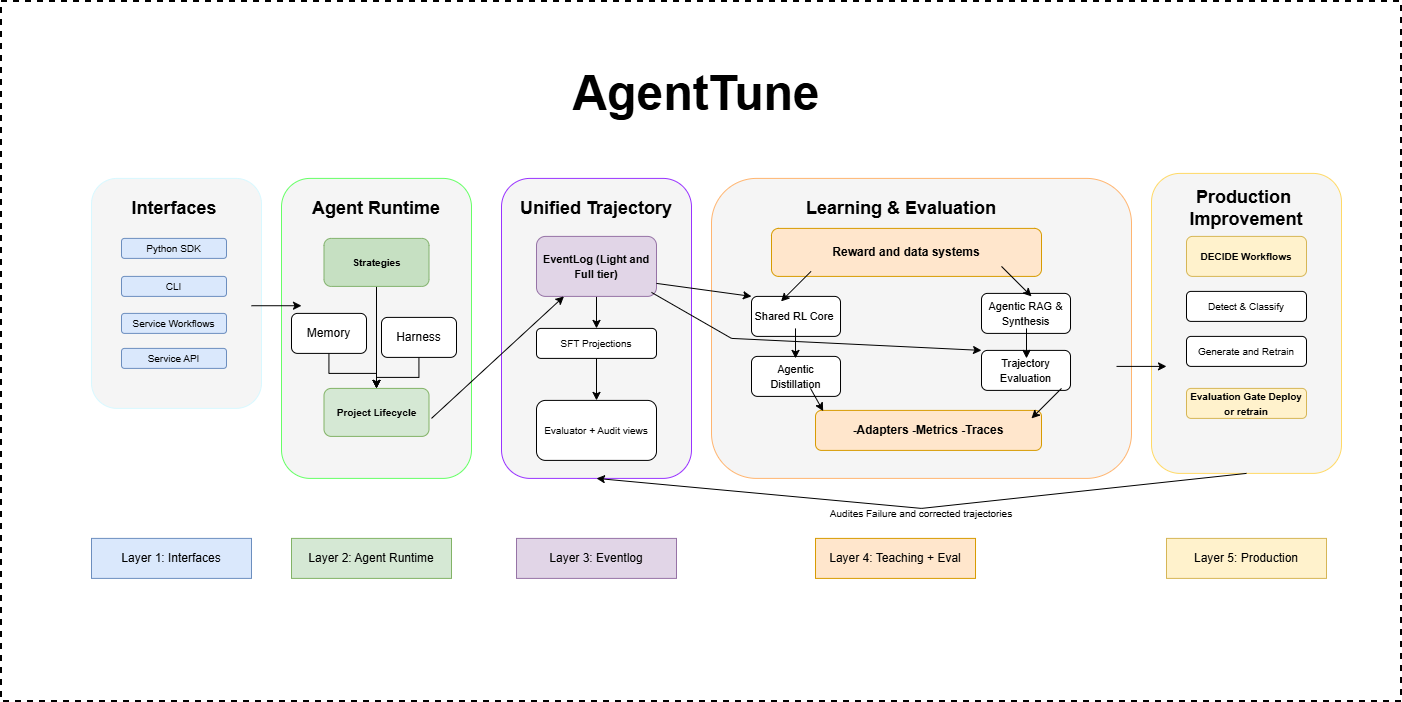}}{\fbox{\parbox[c][0.24\textheight][c]{0.92\textwidth}{\centering Architecture figure\\(figures/architecture.png not supplied)}}}
\caption{\agenttune{} in five layers. Interfaces (1) drive the agent runtime (2), whose episodes normalize into the \eventlog{} (3) in a \code{light} or \code{full} tier; the schema projects out to SFT rows and to the evaluator and audit views, which learning and evaluation (4) consume to produce adapters, metrics, and traces. Production (5) runs DECIDE workflows and the detect--classify--generate--retrain--gate cycle, whose audited failures and corrected trajectories return to Layer~3 as new training data. Agentic distillation is drawn beside the shared RL core rather than inside it because it reaches \code{TRLSFTTrainer} through a caller-supplied \code{trainer\_factory} instead of \code{create\_agentic\_trainer}.}
\label{fig:arch}
\end{figure}

Figure~\ref{fig:arch} lays the library out in five layers, and Table~\ref{tab:packages} maps each layer onto the packages that implement it.

\emph{Interfaces} (Layer~1) are the four ways in: the Python SDK, the \code{agenttune} CLI, the service workflows, and the service API. \emph{Agent runtime} (Layer~2) is where an agent is built and run --- strategies, memory, and a harness, driven through the \code{Project} lifecycle. \emph{Unified trajectory} (Layer~3) is the contract the rest of the library is written against: one \eventlog{} in a \code{light} or \code{full} tier, projected out to SFT rows and to the evaluator and audit views. \emph{Learning and evaluation} (Layer~4) consumes those projections --- the shared RL core, agentic distillation, agentic RAG and synthesis, and trajectory evaluation, over a common base of reward and data systems, and emitting adapters, metrics, and traces. \emph{Production and improvement} (Layer~5) runs DECIDE workflows and closes the loop: detect and classify a failure, generate a correction and retrain, then gate the result before deploying or rolling back. The gate's output returns to Layer~3 as audited failures and corrected trajectories, which is the only cycle in the diagram.

Two details in the figure are worth stating outright, because they are the ones a reader is most likely to assume wrongly. The shared RL core in Layer~4 is reached directly by \code{agenttune.api.train\_agentic}, the RAG training script, \code{AgentTuneGraph}, and DECIDE's retraining bridge; agentic distillation sits beside it rather than inside it, because \code{Project.train(fmt='sft')} and \code{Project.distill()} run supervised fine-tuning through a caller-supplied \code{trainer\_factory} and never call \code{create\_agentic\_trainer} (\S\ref{sec:rlcore},~\S\ref{sec:distill}). And the reward and data systems drawn across the top of Layer~4 --- the rollout engines and the reward registry --- are grouped there by role; both are implemented under \code{agenttune.agentic}.

\begin{table}[tb]
\centering
\scriptsize
\setlength{\tabcolsep}{3pt}
\begin{tabular}{@{}p{2.5cm}p{2.05cm}>{\raggedright\arraybackslash}p{8.2cm}p{1.9cm}@{}}
\toprule
Layer & Package & Contents & Described in \\
\midrule
1 Interfaces & \code{cli/} & the \code{agenttune} console commands and the \code{decide} sub-app; the Python SDK and the \code{[service]} FastAPI app are the other three entry points & \S\ref{sec:packaging} \\
\addlinespace
2 Agent runtime & \code{agentic/} & five strategies, three memory backends and \code{TrajectoryStore}, two harnesses, and the \code{Project} lifecycle & \S\ref{sec:spine} \\
\addlinespace
3 Unified trajectory & \code{agentic/} \code{events.py} & the \eventlog{} schema, its two tiers, and the six projections in and out & \S\ref{sec:eventlog} \\
\addlinespace
4 Learning and evaluation & \code{core/}, \code{backends/}, \code{rag/}, \code{langgraph/}, \code{eval/} & \code{create\_agentic\_trainer} and the five \code{TrlAgentic*} wrappers; three \code{SearchBackend} implementations and the six-stage synthesis pipeline; \code{AgentTuneGraph}; \code{BaseEvaluator}/\code{RLEvaluator}, \code{TrajectoryEvaluator}, the \code{lm-eval} runner and \code{run\_eval}; rollout engines and the 26-entry reward registry & \S\ref{sec:rlcore}--\S\ref{sec:rag}, \S\ref{sec:multiagent},~\S\ref{sec:evalimpl} \\
\addlinespace
5 Production and improvement & \code{decide/} & \code{GraphRunner}, the eight stage types, 24 templates, the append-only audit log, and \code{closed\_loop/} & \S\ref{sec:decide},~\S\ref{sec:selfheal} \\
\midrule
--- & \code{data/}, \code{utils/}, \code{scenarios/} & \code{DataManager}, device and environment diagnostics, checkpoint helpers, scripted domain scenarios & out of scope (\S\ref{sec:scope}) \\
\bottomrule
\end{tabular}
\caption{The five layers of Figure~\ref{fig:arch}, the packages of \code{src/agenttune/} that implement each, and the subsection describing them.}
\label{tab:packages}
\end{table}

\subsection{The \eventlog{} schema}
\label{sec:eventlog}

The schema is one standard-library module (\code{agenttune.agentic.events}). An \eventlog{} is an ordered list of \code{Event}s plus a \code{tier}. Each \code{Event} carries an \code{EventKind}, a free-form \code{payload} dict, and optional \code{token\_span}, \code{logprobs}, and \code{scope} fields. Eight kinds cover the whole trajectory: model text and reasoning, a tool call and its result, an environment observation, a turn boundary, a reward, and a memory operation. \code{REWARD} is the only kind with a meaningful \code{scope}, which is one of step, turn, or episode. Table~\ref{tab:eventkinds} in Appendix~\ref{app:eventkinds} names all eight. Six projections move a trajectory in and out of the schema (Table~\ref{tab:projections}).

\subsubsection{Two tiers}
A \code{light}-tier log is observational: it carries the content of a trajectory (what tool was called, what came back, what the reward was) and is produced by a harness-driven episode, a DECIDE run, or an evaluation trace. A \code{full}-tier log also carries token spans and logprobs and comes only from a rollout engine. Two operations, \code{masked\_tokens()} and \code{as\_dataset\_rows("sft")}, require the \code{full} tier and raise \code{ValueError} on a \code{light}-tier log. \code{masked\_tokens()} returns the token spans the agent itself generated --- the completion spans a trainer computes loss \emph{on}, not the environment spans it masks out; everything else (\code{rewards()}, \code{to\_eval\_dict()}, \code{to\_audit\_records()}, iteration) works on either.

\begin{table}[tb]
\centering
\small
\begin{tabular}{lll}
\toprule
Projection & Direction & Tier produced / required \\
\midrule
\code{from\_trajectory}     & RL rollout $\rightarrow$ \eventlog{}    & \code{full} \\
\code{from\_pipeline\_state} & DECIDE state $\rightarrow$ \eventlog{}   & \code{light} \\
\code{from\_eval\_dict}      & eval dict $\rightarrow$ \eventlog{}      & \code{light} \\
\midrule
\code{as\_dataset\_rows("sft")} & \eventlog{} $\rightarrow$ SFT rows   & requires \code{full} \\
\code{to\_eval\_dict}           & \eventlog{} $\rightarrow$ eval dict  & either \\
\code{to\_audit\_records}       & \eventlog{} $\rightarrow$ audit log  & either \\
\bottomrule
\end{tabular}
\caption{The six \eventlog{} projections. The three \code{from\_*} classmethods only read their input; none of the native formats is modified.}
\label{tab:projections}
\end{table}

\subsubsection{Segment-aware, completion-only loss}
\label{sec:sftproj}
\code{as\_dataset\_rows("sft")} turns one \code{full}-tier log into exactly one training row: a chat-message list plus a per-message weight vector and a derived boolean loss mask (Table~\ref{tab:weights}). Reasoning and final text get normal loss. Tool calls get double weight, since a malformed tool call is a harder failure than imprecise prose. Tool results and observations are masked out; the loss should not cover what a tool returned. The formal objective for parameters $\theta$, query $q$, and generated tokens $y$ is
\begin{equation}
\mathcal{L}(\theta) = -\sum_{i \in \mathcal{I}_{\text{agent}}} w_i \, \log \pi_\theta\!\left(y_i \mid y_{<i}, q\right),
\end{equation}
where $\mathcal{I}_{\text{agent}}$ indexes tokens the agent itself generated and $w_i \in \{0, 1, 2\}$ is the segment weight of the message containing token $i$; environment tokens have $w_i = 0$. Multi-call turns are unwrapped so each call becomes its own weight-2 message with JSON-parsed arguments, matching what the strict inference-time tool-call parser accepts. Appendix~\ref{app:projections} reproduces the projection source.

The projection is lossy in one respect that matters downstream: it has branches for \code{REASONING}/\code{TEXT}, \code{TOOL\_CALL}, \code{TOOL\_RESULT} and \code{OBSERVATION} only. \code{REWARD}, \code{TURN\_COMPLETE} and \code{MEMORY\_OP} fall through without a message, so turn boundaries flatten and a \code{MemoryReActStrategy} trajectory --- the strategy introduced in \S\ref{sec:strategies} precisely because it emits \code{MEMORY\_OP} events --- carries none of its memory operations into training. Five of the eight kinds survive the SFT projection; three do not.

\begin{table}[tb]
\centering
\small
\begin{tabular}{lll}
\toprule
Event kind & Chat role & Segment weight \\
\midrule
\code{REASONING} / \code{TEXT} & \code{assistant} & $1.0$ \\
\code{TOOL\_CALL}              & \code{assistant} & $2.0$ \\
\code{TOOL\_RESULT}            & \code{tool}      & $0.0$ \\
\code{OBSERVATION}             & \code{user}      & $0.0$ \\
\bottomrule
\end{tabular}
\caption{Segment weights emitted by \code{as\_dataset\_rows("sft")}; \code{loss\_mask} is \code{[w == 0.0 for w]}. Note the polarity: unlike \code{attention\_mask} or TRL's \code{completion\_mask}, \code{True} here means \emph{excluded} from the loss.}
\label{tab:weights}
\end{table}

\subsubsection{Projections and consumers}
\code{to\_eval\_dict()} produces the tool-call and tool-output fields consumed by the trajectory evaluator. \code{to\_audit\_records()} produces the trajectory identifier, stage name, and state snapshot read by the closed loop's failure detector. A tool name and its arguments are folded into one stage name, so loop collapse is triggered only by an identical repeated action, not by two calls to the same tool with different arguments. \code{Project.heal()} applies this projection to trajectories collected by the spine.

\subsubsection{Schema properties and their limits}
\label{sec:schemaprops}

Four properties are what make one record serve five stages, and stating them makes it testable whether an implementation holds them.
\emph{Order preservation}: an \eventlog{} is a totally ordered sequence, and every projection must visit events in that order, because a trajectory's meaning is its sequence.
\emph{Tier monotonicity}: a \code{full}-tier log carries everything a \code{light}-tier log carries plus token spans and logprobs, so any consumer of \code{light} accepts \code{full} unchanged.
\emph{Weight partition}: every event that reaches a training row is assigned exactly one segment weight, so agent-generated and environment-supplied tokens partition the sequence rather than overlapping.
\emph{Read-only sources}: the three \code{from\_*} classmethods copy out of their native format and never mutate it, so adopting the schema cannot disturb an existing pipeline.

Two of the four are conditional in the current release. Weight partition holds within the SFT projection but not across the schema, because the three kinds that projection drops (\S\ref{sec:sftproj}) receive no weight at all. Tier monotonicity holds for every consumer except the two operations that require \code{full} and raise on \code{light}, which is the intended asymmetry rather than a violation. Order preservation and read-only sources hold unconditionally, and the test suite covers both.

The schema is also deliberately narrow, and five things it cannot represent bound where it is the right abstraction. \emph{Concurrent tool calls} serialize: a turn that dispatches three tools in parallel is recorded as three ordered \code{TOOL\_CALL} events, so wall-clock structure is lost even though causal order is kept. \emph{Streaming} has no representation; an event is recorded once complete, so partial-token supervision is out of reach. \emph{Human turns} have no kind of their own and enter as \code{OBSERVATION}, which makes a human correction indistinguishable from an environment response at training time. \emph{Per-agent attribution} is absent in multi-agent rollouts, where several policies contribute to one log with no field naming the author of an event, so credit assignment across agents must be reconstructed externally. And \emph{branch points} flatten: the search that \code{TreeOfThoughtsStrategy} performs and the retry that \code{ReflexionStrategy} performs are both linearized into the accepted path, so the rejected branches, which are exactly the negative examples a preference method wants, are not recoverable from the log. Extending the schema to record them is the most substantial open item in its design.

\subsection{The agentic spine}
\label{sec:spine}

The spine is the build-and-run layer: an \code{AgentStrategy} plus a \code{Harness}, driven by a \code{Project} that logs every step to the \eventlog{}.

\subsubsection{Agent strategies}
\label{sec:strategies}

Five control policies sit behind one four-method interface (\code{init}, \code{propose}, \code{observe}, \code{is\_done}). Every strategy keeps model access behind an injected callable, so a scripted function stands in for a model in tests and a real \code{model.generate(\dots)} call drops in unchanged.

\begin{itemize}[noitemsep,topsep=2pt]
\item \code{ReActStrategy} \citep{yao2023react} proposes one action per step and decides when to stop.
\item \code{PlanExecuteStrategy} \citep{wang2023planandsolve} calls the policy once for a complete plan, then replays the actions through \code{propose}.
\item \code{ReflexionStrategy} \citep{shinn2023reflexion} retries a task up to \code{max\_attempts} times and carries a verbal self-critique from each failed attempt into the next. It is driven by \code{run\_reflexion} rather than a single \code{run\_episode}.
\item \code{MemoryReActStrategy} augments ReAct with persistent memory. \code{init} recalls the $k$ most relevant items, while each \code{observe} call stores the step's observation as an episodic \code{MemoryItem} and emits a \code{MEMORY\_OP} event.
\item \code{TreeOfThoughtsStrategy} \citep{yao2023tot} proposes and scores several actions per step. It can select the leading candidate greedily through \code{run\_episode} or backtrack over harness snapshots through \code{.search()}.
\end{itemize}

\code{run\_episode(strategy, harness, task)} is the driver every strategy plugs into unchanged: it calls \code{strategy.init}, then loops \code{propose}\,$\to$\,\code{harness.step}\,$\to$\,\code{observe} until \code{is\_done}, and returns \code{harness.to\_eventlog()}. \code{AgentState} is a plain dataclass (\code{task}, \code{events}, \code{step}, \code{done}, \code{scratch}), where \code{scratch} is the strategy's private working area (the tool set for ReAct, the plan and cursor for Plan-and-Solve, the search frontier for ToT).

\subsubsection{Memory}
\label{sec:memory}

Three backends implement one \code{BaseMemory} interface: \code{InContextMemory} (recent-$k$), \code{VectorMemory} (semantic recall via a sentence-transformer embedder), and \code{GraphMemory} (relational and temporal). A \code{TrajectoryStore} persists past rollouts for replay and training. A \code{Scope} (\code{agent\_id}, \code{namespace}) determines whether memory is shared across agents or private to one.

\subsubsection{Harnesses and conformance}
\label{sec:harness}

\code{Harness} is a Gym-style abstract base class with reset, step, and action-space methods. A separate \code{Harness\-Capabilities} object declares snapshot, streaming, tool-boundary interruption, stepwise-turn, and maximum-step behavior.

\begin{itemize}[noitemsep,topsep=2pt]
\item \code{DictToolHarness} runs plain Python callables in process without a sandbox. An unknown tool name or raised exception becomes a \code{TOOL\_RESULT} containing \code{"error: \dots"} and a step reward of $-1.0$, rather than crashing the harness. Snapshot and restore operations use an in-process pickle blob.
\item \code{OpenEnvHarness} wraps an OpenEnv \citep{openenv2025} \code{Environment}, a test fake, or any object that exposes \code{reset()} and \code{step(action)}. Tool calls execute in an isolated remote sandbox.
\end{itemize}

\code{run\_conformance(harness)} exercises the capability flags rather than trusting them, snapshotting, stepping, restoring, and checking the event log reverted, and returns a \code{ConformanceReport(passed, drift)}. A harness that claims \code{supports\_snapshot=True} but whose \code{restore()} does not revert state shows up as drift; this is a real regression check. Appendix~\ref{app:harness} shows a run and the event sequence it produces.

\subsubsection{The Project lifecycle}
\label{sec:project}

\code{Project(strategy, harness)} exposes the lifecycle as \code{infer}, \code{collect}/\code{collect\_rollout}, \code{evaluate}/\code{evaluate\_agentic}, \code{train}, \code{distill}, and \code{heal}. Collection through a real rollout engine produces \code{full}-tier logs; evaluation combines task success with optional trajectory metrics; training and distillation consume the resulting projections; and healing converts failures back into training data. Each call emits a \code{LifecycleEvent}, making the sequence inspectable without coupling the \code{Project} to a particular trainer or model.

\subsection{The shared RL core}
\label{sec:rlcore}

The RL core is the only place reinforcement-learning weights change. The public training wrapper \code{agenttune.api.train\_agentic}, RAG training, multi-agent orchestration and DECIDE's retraining step all call \code{create\_agentic\_trainer} directly rather than carrying a trainer each, so an algorithm added once is available to all of them. The spine is the deliberate exception. \code{Project.train} and \code{Project.distill} take a caller-supplied \code{trainer\_factory} and drive supervised fine-tuning over \code{as\_dataset\_rows("sft")} output (\S\ref{sec:distill}); they never construct a trainer themselves, which keeps \code{Project} importable and testable with no GPU and no training stack loaded.

\subsubsection{The factory API}
\code{create\_agentic\_trainer(algorithm, **kwargs)} lower-cases the requested algorithm name, looks it up in a five-member enum, and instantiates the matching \code{TrlAgentic*} wrapper with every kwarg forwarded. Each wrapper calls \code{inspect.signature()} on the installed TRL's \code{*Config} and \code{*Trainer} classes to split kwargs between them, so the accepted parameter surface tracks the installed TRL version instead of a frozen \agenttune{}-side schema. Passing \code{tools=[\dots]} sets agentic mode: the wrapper builds a multi-turn tool-calling rollout function via \code{create\_rollout\_fn} and monkey-patches TRL's generation path to call it, since the installed TRL cannot run a \code{BaseTool} object natively. Table~\ref{tab:algos} summarizes the five.

A \code{backend} argument selects the trainer family: \code{"trl"} (the default, imported eagerly and used for every example here) or \code{"unsloth"} \citep{unsloth} (a separate \code{pip install unsloth}, for faster low-memory training). Unsloth patches \code{transformers}/\code{trl}/\code{peft} at import time, so its trainers load only when a caller passes \code{backend="unsloth"}; \code{PURE\_TRL\_MODE=1} forces \code{import unsloth} to never run. The same \code{tools}, \code{train\_dataset}, and \code{reward\_funcs} arguments carry across both backends.

\begin{table}[tb]
\centering
\scriptsize
\setlength{\tabcolsep}{3pt}
\begin{tabular}{@{}p{3.15cm}p{4.0cm}p{5.1cm}@{}}
\toprule
Algorithm & Needs & Rollout modes \\
\midrule
GRPO \citep{shao2024deepseekmath}  & reward function(s) & agentic tool loop; \code{combine\_rewards} applied \\
DPO \citep{rafailov2023dpo}   & \code{chosen}/\code{rejected}, \emph{or} a reward fn & offline; reward-ranked; agentic \\
PPO \citep{schulman2017ppo}   & reward fn/model \emph{and} value fn/model & agentic; patches \code{batch\_generation} \\
RLOO \citep{ahmadian2024rloo}  & reward function & agentic; reward fn forwarded as-is \\
BCO \citep{jung2024bco}   & \code{completion}/\code{label}, \emph{or} a reward fn & offline; online (regenerates per epoch) \\
\bottomrule
\end{tabular}
\caption{The five dispatchable algorithms behind \code{create\_agentic\_trainer}. GRPO is the default for agentic training. Only GRPO routes \code{reward\_funcs} through the weighted \code{combine\_rewards} combinator.}
\label{tab:algos}
\end{table}

\begin{lstlisting}[caption={Training a tool-using agent with GRPO on Qwen2.5-1.5B-Instruct \citep{yang2024qwen25technicalreport}. \code{tools} are plain callables; \agenttune{} builds the model-facing JSON schema from the type hints and docstring. Extra dataset columns (here \code{answer}) are forwarded to the reward functions as kwargs.}, label={lst:grpo}]
from datasets import Dataset
from agenttune.core.backend_factory import create_agentic_trainer

def calculator(expression: str) -> str:
    """Evaluate an arithmetic expression and return the result.

    Args:
        expression: e.g. "12 * (3 + 4)".
    """
    ...  # safe AST eval

def correct_answer_reward(completions, answer=None, **kwargs):
    golds = answer or [None] * len(completions)
    return [1.0 if g is not None and str(g) in str(c) else 0.0
            for c, g in zip(completions, golds)]

def used_calculator_reward(completions, **kwargs):
    return [0.2 if "<tool_call>" in str(c) else 0.0 for c in completions]

train_dataset = Dataset.from_dict({
    "prompt": ["What is 12 * (3 + 4)?", "What is 100 / 4 - 5?"],
    "answer": ["84", "20"],
})

trainer = create_agentic_trainer(
    algorithm="grpo",
    model="Qwen/Qwen2.5-1.5B-Instruct",
    tools=[calculator],
    reward_funcs=[correct_answer_reward, used_calculator_reward],
    reward_weights=[0.8, 0.2],
    train_dataset=train_dataset,
    system_prompt="You are a calculator agent. Use the tool, then answer.",
    num_generations=4, per_device_train_batch_size=4,
    gradient_accumulation_steps=1,
    max_steps_per_turn=6,   # tool-call loop depth per rollout
    max_steps=30,           # total optimizer steps
    peft_config={"r": 16, "lora_alpha": 32, "task_type": "CAUSAL_LM",
                 "target_modules": ["q_proj", "k_proj", "v_proj", "o_proj"]},
    output_dir="./runs/grpo-calculator",
)
results = trainer.train()
\end{lstlisting}

\subsubsection{Rollout engines}
Three interchangeable engines back generation: \code{transformers} (local, single process), \code{vllm} \citep{kwon2023vllm} (batched, colocate or server mode), and \code{api} (any LiteLLM \citep{litellm} provider, one class speaking OpenAI, Anthropic, Groq, OpenRouter, Ollama, Together, and Azure, keyed off the model-string prefix). Each is usable standalone; \code{create\_rollout\_fn(\dots)(prompts)} returns a dict carrying the trajectories, the final responses, the rewards, the token ids and logprobs, and the per-rollout tool-call counts; Table~\ref{tab:rolloutkeys} in Appendix~\ref{app:rollout} gives the full contract key by key. The rollout engine and the trained model are separate, so a run can generate rollouts with a hosted 70B model and compute gradients against a local 1.5B one.

\subsubsection{Rewards and the combinator}
\code{REWARD\_REGISTRY} holds 26 named reward functions (Table~\ref{tab:rewards}), correctness, format/structure, tool-use, retrieval-grounding, numeric-match, and task-specific variants for SQL and FinQA \citep{finqa}, each returning \code{list[float]}, one score per completion; Appendix~\ref{app:rewards} catalogs all 26. \code{combine\_rewards(reward\_funcs, weights)} accepts callables or registry names, renormalizes weights to sum to one, and returns one GRPO-compatible callable. A reward function that fails under both calling conventions raises rather than scoring \code{0.0}; training against a dead signal is worse than a crash.

\subsubsection{LLM judge and reward-hacking defenses}
\code{LLMJudge} runs as a hosted API judge (provider inferred from the model name), a local \code{transformers} model, or a local \code{vllm} model, or reuses an already-loaded engine to avoid a second model load. \code{evaluate\_batch(\dots, mode="relative")} scores all trajectories in one prompt and enforces a strict ranking (better signal for GRPO); \code{mode="absolute"} scores each independently. Scores are cached by \code{(task, trajectory, criteria)}. The judge is aimed at held-out evaluation reports, groundedness in particular, rather than the training loop: an API call per rollout inside GRPO trades a large latency cost for a small signal gain. For an in-loop learned scorer, \code{distilled\_judge\_reward} runs a compact \code{text-classification} model over the prompt and completion instead.

Three defenses are built in: score clamping when a policy over-retrieves; judge-prompt rotation (a random draw per call, so a policy cannot fit one fixed prompt); and relative batch scoring. A separate \code{RuleGuardCombinator} deterministically clamps any judge score to $\le 0.2$ when the completion shares no vocabulary with the retrieved chunks, a hard guard against an LLM judge rewarding an ungrounded hallucination.

\subsubsection{Algorithms and configuration}
The five wrappers share the same public surface for model, dataset, PEFT, system prompt, rollout, and output configuration while preserving their underlying TRL semantics. GRPO and RLOO use reward functions; DPO and BCO consume preference/labeled data or can derive it from rollouts; PPO additionally requires value and reward components. The important distinction is that AgentTune replaces completion-only generation with the same multi-turn tool loop used by the agentic runtime, while leaving the underlying optimization algorithms unchanged. Appendix~\ref{app:rollout} records the full rollout contract and configuration surface.

\subsection{Agentic distillation}
\label{sec:distill}

\code{Project.distill(student, trainer\_factory=\dots)} formulates distillation as \emph{agentic behavior cloning}. A teacher agent produces \code{full}-tier trajectories; those with ungrounded citations, schema errors, or wrong final answers are filtered out; the survivors become SFT rows via \code{as\_dataset\_rows("sft")}, so the segment-aware completion-only loss of \S\ref{sec:eventlog} applies automatically. The student reads the retrieved context to condition its decisions, but the loss covers only its own reasoning, tool calls, and answers. \code{Project} delegates the actual training to a caller-supplied \code{trainer\_factory} so it stays GPU- and dependency-free; the factory typically wraps TRL's \code{SFTTrainer} with LoRA \citep{hu2022lora} and \code{completion\_only\_loss=True} (Listing~\ref{lst:distill}). \code{Project.train(fmt='sft')} rides identical rails; the only difference is whose trajectories and whose weights.

\begin{lstlisting}[caption={Collect teacher trajectories and distill into a sub-3B student. \code{sft\_dataset()} previews the exact rows the trainer will see, GPU-free.}, label={lst:distill}]
from agenttune.agentic import Project
from agenttune.agentic.rollout_engines.rollout_factory import create_rollout_engine

teacher_engine = create_rollout_engine(
    backend="transformers", model_path="Qwen/Qwen2.5-7B-Instruct")

teacher = Project()
logs = teacher.collect_rollout(teacher_engine, tasks, tools=tools, max_steps=6)

rows = teacher.sft_dataset()   # list[{"messages","segment_weights","loss_mask"}]

result = teacher.distill(
    student="Qwen/Qwen2.5-0.5B-Instruct",
    trainer_factory=RealSFTTrainerFactory(out_dir="./student"),
)
\end{lstlisting}

Three ways get \code{full}-tier trajectories onto a \code{Project}, in increasing realism.

\begin{itemize}[noitemsep,topsep=2pt]
\item \code{add\_trajectory} takes hand-built \code{EventLog(tier="full")} objects, which suit a fixed demonstration set that teaches one specific output contract.
\item \code{collect\_rollout(engine, tasks)} runs a real teacher model, text-only.
\item \code{collect\_rollout(engine, tasks, tools=[\dots])} runs an agentic teacher, whose real \code{TOOL\_CALL}/\code{TOOL\_RESULT} pairs become \code{<tool\_call>\{\dots\}</tool\_call>} training targets, so the student learns the teacher's tool-use policy and not only its prose.
\end{itemize}
 \code{teacher.sft\_dataset()} previews the exact rows GPU-free; \code{distill()} raises a clear \code{ValueError} rather than training on nothing if no trajectories were collected; and every stage emits \code{("distill", "started"/"done", \{\dots\})} lifecycle events carrying \code{n\_rows} and \code{n\_teacher\_trajectories} so a caller can confirm the run saw what it expected before waiting on a GPU job. The entire path up to \code{.train()} is exercisable with zero GPU and zero API key using a \code{DemoRolloutEngine} teacher and a stub \code{trainer\_factory}, which still routes through the same \code{collect\_rollout}\,$\to$\,\code{sft\_dataset}\,$\to$\,\code{distill} machinery.

\subsection{Agentic RAG and data synthesis}
\label{sec:rag}

\code{agenttune.rag} holds two things that are usually kept apart: the machinery for training a search policy against a real retrieval backend, and a pipeline that turns a raw corpus into the grounded multi-hop questions such a policy needs. Both write \eventlog{}s, so a run of either feeds the stages of \S\ref{sec:rlcore} and \S\ref{sec:evalimpl} unchanged.

\subsubsection{Retrieval backends and search training}

\code{agenttune.rag} combines a retrieval backend, a \code{SearchCorpusTool}, and search and correctness rewards in a GRPO trainer, so the policy learns when to retrieve rather than consuming a fixed retrieve-then-read context \citep{lewis2021retrievalaugmentedgenerationknowledgeintensivenlp,ram-etal-2023-context}. Its three implementations of the \code{SearchBackend} protocol are \code{SQLiteFTSBackend}, which uses FTS5 and BM25 ranking \citep{robertson2009bm25}; \code{ChromaBackend} \citep{chroma}, which provides dense retrieval; and \code{HybridBackend}, which fuses BM25 and BGE-M3 \citep{bgem3} results through reciprocal rank fusion \citep{cormack2009rrf}. On the multi-hop CUAD \citep{hendrycks2021cuad} questions, their gold-chunk R@8 values are 19.3\%, 18.1\%, and 21.3\%, respectively; the corresponding gold-passage hit rates at top 8 are 83.3\%, 81.7\%, and 90.0\%. The backends are pickle-safe because their constructors retain file paths rather than live connections, allowing rollout workers to run in separate processes. \code{SearchCorpusTool} is a \code{BaseTool} instance and can be passed directly to \code{create\_rollout\_fn(tools=[\dots])}.

\subsubsection{The six-stage synthesis pipeline}

The synthesis entry point, \code{rag.synthesis.run\_pipeline}, turns a document corpus into a grounded, difficulty-tagged multi-hop QA dataset. It reuses primitives from GRADE \citep{grade2024}, RAGAS \citep{es2024ragas}, and MHTS \citep{mhts2025} in six stages.

\begin{enumerate}[topsep=2pt,itemsep=1pt,label=\arabic*.,leftmargin=1.8em]
\setcounter{enumi}{-1}
\item The corpus is chunked and converted into a typed graph after an LLM extracts entities, keyphrases, and a summary from each chunk. Edges are marked as \code{exact}, \code{contextual}, or \code{abstract}.
\item Bounded simple paths of two to five hops are sampled with \code{networkx} \citep{hagberg2008networkx}; a guard prevents any edge from crossing a document boundary.
\item Answer-first generation \citep{mhts2025} fixes the answer in an early passage and asks for a question that requires the remaining evidence chain. Each path uses one structured JSON-mode model call.
\item Closed-loop verification applies answerability, chain-dependency, and retrieval-necessity checks. These follow \citet{min}, who showed that compositional multi-hop questions are frequently answerable from a single passage, so a question is kept only when the second hop is required. A failed candidate receives one targeted regeneration attempt before rejection.
\item A two-dimensional matrix combines hop count with embedding-based retrieval difficulty, computed with a power mean at $p=-3$, and supports balanced resampling.
\item The final stage reports answerability, faithfulness, diversity, and coverage, then creates a leakage-safe train/validation split through shared-gold-chunk union--find components and greedy bin packing. The final answerability gate used in dataset curation is applied after the three verification checks.
\end{enumerate}

Both the LLM and the embedder are injected \code{Protocol}s, so the whole pipeline runs on CPU with no network via \code{FakeLLMClient}/\code{FakeEmbedder}; production runs swap in \code{GroqLLMClient} / \code{OpenAICompatLLMClient} and \code{BGEM3Embedder}. Every stage writes CSV and Parquet with per-LLM-call token, cost, and latency metadata; only Stage~0 (the expensive, corpus-deterministic one) is cached. The run also emits \code{dataset\_grpo.jsonl} (prompt / gold answer / gold path, ready for GRPO), \code{dataset\_all.csv} (every candidate including rejected ones with a \code{discard\_reason}), \code{metrics.json}, and \code{manifest.json}.

\subsubsection{Standalone utilities}

Several pieces of the RAG package stand alone: \code{split\_by\_gold\_chunks} (a leakage-safe train/val splitter over any objects with \code{gold\_chunk\_ids}; union-find components plus greedy bin-packing, returns \code{(train, val, report)} and never violates the disjointness guarantee to hit a requested fraction, marking the result \code{degenerate} instead); \code{extract\_json} (the direct-parse\,$\to$\,fenced-block\,$\to$\,brace-scan extractor every synthesis stage uses for messy LLM JSON); \code{build\_curriculum}/\code{CurriculumSampler} (an easy$\rightarrow$hard training curriculum combining a solve-difficulty probe and a retrieval-difficulty signal); \code{TrajectoryStore} (a stdlib-\code{sqlite3}, pickle-safe log of every rollout, question, reward components, per-step thoughts/actions/observations, with \code{get\_by\_reward\_range} for dataset building and a ready-made \code{on\_trajectory\_end} callback); a MEM1-style \citep{mem12025} running-state compressor that keeps rollout context roughly constant regardless of how many searches an agent runs; an XML tool-call parser for models whose chat template emits \code{<function=\dots>} instead of JSON; and a bring-your-own-documents CLI (\code{python -m agenttune.rag.scripts.build\_index\_from\_docs}) that ingests mixed-format files (text/Markdown/CSV/JSON directly, PDF/DOCX/PPTX/HTML with one optional library each) into either backend.

\subsection{DECIDE: decision workflows as YAML}
\label{sec:decide}

\code{GraphRunner.from\_template(\dots)} compiles a YAML template into a LangGraph \code{StateGraph}, one node per stage (Table~\ref{tab:stages}). Four classes do the work:

\begin{itemize}[noitemsep,topsep=2pt]
\item \code{ConfigLoader} resolves the template's \code{extends:} chain, deep-merges the global config, auto-generates the edge list, and rejects any template identifier containing \code{..}.
\item \code{TemplateRegistry} indexes a directory of templates by identifier.
\item \code{GraphRunner} compiles a template and executes it.
\item \code{PipelineState} threads the working state through every stage: each stage's outputs, the step history, the running verdict with its label, confidence and reason, and the completion and next-stage flags.
\end{itemize}

Step-limit enforcement and an append-only audit log are built in; 24 templates ship (12 BFSI: KYC, fraud, loan approval, sanctions, card decline, claims triage, \dots; 10 generic: classification, sentiment, summarization, entity extraction, multi-judge scoring, \dots; 2 custom). \code{\{input\_text\}} and \code{\{stage\_id.output.field\}} interpolation lets a stage's prompt or a \code{tool\_call}'s arguments reach the raw input or an upstream stage's parsed output. Appendix~\ref{app:config} excerpts the environment config every template extends.

\begin{table}[tb]
\centering
\scriptsize
\setlength{\tabcolsep}{3pt}
\begin{tabular}{lp{5.0cm}}
\toprule
Stage type & What it does \\
\midrule
\code{llm\_call}     & one model call, reply validated against an \code{output\_schema} \\
\code{router}        & LLM-based (has a \code{model:}) or rules-based (no \code{model:}) dispatch to one of several \code{on\_result} targets, through one stage type \\
\code{llm\_judge}     & scores an upstream output on a rubric \\
\code{rules}         & deterministic gate: a list of \code{condition} strings over \code{stage\_outputs}, evaluated by a hand-written AST walker (only \code{== != < <= > >= and or not}, dotted lookups, JSON literals; no \code{eval()}, no function calls); first failing rule short-circuits with \code{goto}\,+\,\code{inject} \\
\code{parallel}      & fan-out to independent branches, run concurrently via \code{asyncio.gather}, merged back into \code{stage\_outputs} by branch id \\
\code{tool\_call}     & invoke a tool from the same registry the spine uses, with interpolated args and a per-stage timeout \\
\code{human\_review}  & filesystem handshake: write \code{pending\_reviews/\{id\}.json}, poll for a decision file until \code{timeout\_seconds}. Fails closed: a timeout yields the decision \code{"timeout"} and routes to \code{on\_failed}, then \code{on\_denied} \\
\code{output}        & finalize \code{verdict}/\code{confidence}; destinations are dispatched by \code{DestinationRouter} after the run completes, not by the stage \\
\bottomrule
\end{tabular}
\caption{The eight DECIDE stage types, all registered and executable. A straight \code{llm\_call}\,$\to$\,\code{output} chain is the easy case; real pipelines branch with \code{rules} and \code{router}.}
\label{tab:stages}
\end{table}

The compiled graph can be run for inference, collection, evaluation, or training. Collection records training-shaped examples from the in-memory pipeline state, while evaluation can enforce configured thresholds before a training or deployment step. A bundled BFSI template illustrates the YAML-to-graph path; Appendix~\ref{sec:examples} gives the complete invocation.

\subsection{The self-healing closed loop}
\label{sec:selfheal}

\code{agenttune.decide.closed\_loop} reads the audit log (or spine-collected trajectories, via \code{Project.heal()}) and runs two stages connected by a thread-safe FIFO-capped \code{TrainingBuffer}.

\emph{Stage 1 turns a failure into an example.} \code{FailureDetector} flags a run that falls below a judge threshold or passes a revisit limit. \code{FailureClassifier}, an LLM call, assigns one of five root causes: wrong tool, wrong routing, incomplete reasoning, hallucinated output, or loop collapse. \code{TrainingExampleGenerator} then synthesizes a corrected response, validates it by replay, and emits a \code{TrainingExample} --- a multi-completion form for GRPO, or a \code{chosen}/\code{rejected} pair for DPO and BCO.

\emph{Stage 2 decides whether to retrain and whether to ship.} \code{RetrainingTrigger} evaluates six conditions (Table~\ref{tab:triggers}) and two safety gates before atomically draining the buffer; the gates prevent a double-fire while a retrain is already running, and refuse to train on a mostly-bad buffer. \code{BackgroundRetrainer} then runs the retrain on a daemon thread, so the loop keeps ingesting while it trains. Finally \code{DeploymentGate.evaluate\_decision} scores old against new on a held-out set reconstructed from past \emph{successful} runs, and separately checks that the trajectory-eval score has not regressed; only if both pass does \code{apply\_decision} deploy the new adapter, with an automatic backup and a rollback path.

\begin{table}[tb]
\centering
\scriptsize
\setlength{\tabcolsep}{3pt}
\begin{tabular}{llp{3.6cm}}
\toprule
\# & Name & Fires when \\
\midrule
T1 & \code{total\_failures\_exceeded} & buffer size $\ge$ threshold \\
T2 & \code{dominance:<type>}          & one \code{root\_cause} $\ge$ the dominance ratio of the buffer \\
T3 & \code{enough\_examples}          & accepted-example count $\ge$ minimum \\
T4 & \code{reward\_drift}             & rolling mean episode reward drifts below baseline \\
T5 & \code{novel\_failure:<type>}     & a never-seen \code{root\_cause} recurs $\ge$ a minimum count \\
T6 & \code{staleness}                 & time since last retrain $\ge$ a window, with a minimum buffer \\
\bottomrule
\end{tabular}
\caption{The six \code{RetrainingTrigger} conditions; any one fires a retrain. Two safety gates run first.}
\label{tab:triggers}
\end{table}

\paragraph{Isolated tool execution.}
\code{IsolatedTool}/\code{isolate\_tools()} run a tool in a separate OS process (Level~1 of a planned isolation ladder) so a crashing, hanging, or untrusted tool cannot take down the rollout loop or read the training process's memory. It presents the identical \code{BaseTool} interface, and if isolation is impossible (an un-picklable tool holding a lock or a socket, a worker crash, a timeout) it falls back to in-process execution and records which path ran in \code{metadata["isolation"]}; \code{require\_isolation=True} turns a silent fallback into an explicit error result for callers that must not run untrusted code in-process.

\subsection{Multi-agent orchestration}
\label{sec:multiagent}

\code{AgentTuneGraph} composes rollout nodes (from \code{create\_rollout\_fn}) and judge nodes (\code{LLMJudge} instances) into one LangGraph \code{StateGraph}, auto-wiring outputs between nodes by inspecting each node's parameter names. It compiles to three drop-in callables: \code{compile\_rollout()} for \code{GRPOTrainer(rollout\_func=\dots)}, \code{compile\_grpo\_reward()} for \code{reward\_funcs=[\dots]} (producing a per-completion \code{judge\_rewards} vector, which is what GRPO needs), and \code{compile\_reward()}, a plain trajectory$\rightarrow$float scorer usable outside training entirely. An optional router picks one rollout node per batch instead of running all in sequence. Every node is duck-typed, so the whole compile-and-run mechanism is testable with trivial fakes.

\subsection{Evaluation}
\label{sec:evalimpl}

\code{agenttune.eval} has four parts.

\begin{itemize}[noitemsep,topsep=2pt]
\item \code{BaseEvaluator} and \code{RLEvaluator} drive a generation loop over a dataset, auto-detecting prompt and target columns, and compute ROUGE \citep{lin2004rouge}, BLEU \citep{papineni2002bleu}, perplexity, KL divergence against a reference model, policy entropy, win-rate, reward-margin, preference accuracy, and, for DPO, implicit-reward metrics that need no reward model.
\item \code{TrajectoryEvaluator} computes deterministic procedural metrics from any \code{to\_eval\_dict()}-shaped log with no LLM (Table~\ref{tab:trajmetrics}); its full \code{evaluate\_batch} adds an LLM-judge rubric for goal completion, grounding and safety on top.
\item \code{LMEvalRunner} shells out to the EleutherAI \code{lm-eval} CLI \citep{eval-harness} for standard benchmarks --- ARC, GSM8K, MMLU, HellaSwag, HumanEval, MBPP and more --- parsing its JSON back into \code{EvalResult} objects.
\item \code{run\_eval}, a self-contained module, scores tool-using agents on email-search, FinQA, and file-ingestion task shapes, with a zero-model path that takes pre-generated completion strings straight to scoring.
\end{itemize}

Adding a metric means subclassing \code{Metric} with a \code{compute(predictions, references) -> dict} method; \code{safe\_compute()} converts a failed batch into an explicit error metric. Code-generation evaluation uses a timed subprocess with restricted builtins and \code{PassAtKMetric}; this is a robustness measure, not a security boundary, because filesystem and OS-level confinement are not enforced.

\begin{table}[tb]
\centering
\scriptsize
\setlength{\tabcolsep}{3pt}
\begin{tabular}{lp{5.2cm}}
\toprule
Metric & Measures \\
\midrule
TAC  & tool-argument correctness, jsonschema-validated against registered schemas \\
ARR  & API-redundancy ratio: duplicate calls / total \\
SCSR & self-correction success rate: recovered from a trailing error \\
RAD  & reasoning-to-action density (character-count proxy) \\
LCF  & loop-collapse frequency: 3+ identical consecutive calls \\
PAS  & plan adherence: planned steps that appear in the trajectory \\
TER  & tool efficacy: unique tool outputs / total outputs \\
PMED / ASE & path edit-distance / action-state efficiency vs.\ a supplied golden trajectory \\
\bottomrule
\end{tabular}
\caption{Deterministic trajectory metrics computed by \code{TrajectoryEvaluator} with no LLM; \code{Project.evaluate\_agentic()} surfaces these.}
\label{tab:trajmetrics}
\end{table}

\subsection{Tool library, CLI, and packaging}
\label{sec:packaging}

Six zero-dependency builtin tools (file read/write/list, Python and Bash execution, grep) import directly with no extras. Six more wrap \code{langchain\_community} toolkits (SQL, web search, GitHub, Slack, Playwright, HTTP). An OpenEnv adapter discovers and wraps remote sandboxed tools over MCP \citep{mcp2024}. Any object with \code{.name} and \code{.to\_schema()}, or a plain callable with a docstring, works as a tool. AgentTune has three interfaces: the Python SDK used throughout this paper, the \code{agenttune} CLI (\code{train}, \code{pipeline}, \code{version}, and the \code{decide} sub-app), and a FastAPI + WebSocket service with an operator UI behind the \code{[service]} extra. \agenttune{} requires a current Python, and PyTorch with a CUDA GPU for training. A single \code{pip install -e .} covers the spine, RL training, the RAG package, extra eval metrics, the PostgreSQL DECIDE destination, and the OpenEnv sandbox; \code{[service]}, \code{[docs]}, \code{[dev]}, and \code{[flash-attn]} are optional extras. An optional, unofficial Unsloth backend is installable separately.

\section{Case Studies}
\label{sec:casestudies}

This section presents representative evidence for the core claims: small training runs, the DECIDE closed loop end to end, and the testability of the orchestration. Section~\ref{sec:usecase} then works one domain, legal contract search, from data synthesis to an on-premises distilled agent. All GPU results reported here and in \S\ref{sec:usecase} were captured on a single NVIDIA RTX PRO 6000 Blackwell Max-Q workstation GPU. One caveat on scope: the LegalBench-RAG transfer evaluation of \S\ref{sec:legal} is an internal result --- neither its harness nor its outputs are present in the public mirror.

\paragraph{Claims and where they are tested.} Three propositions follow from the design, and we state where each is examined, including where the evidence is currently weakest.
\emph{Schema fidelity} --- a single record can serve five stages without a per-boundary adapter. \S\ref{sec:eventlog} gives the six projections and \S\ref{sec:schemaprops} states the four properties they must preserve, together with the two that hold only conditionally today.
\emph{Token ownership} --- keeping segment weights on the trajectory is what makes environment masking a schema property rather than a per-trainer convention. \S\ref{sec:sftproj} defines the weighting; \S\ref{sec:design} reports what the shipped training path consumes, which is less than the schema exports.
\emph{Instrumentation cost} --- one schema replaces the pairwise converters a multi-stage pipeline would otherwise need. Table~\ref{tab:related} states the counting argument, six projections against up to fifteen converters. This is the proposition with the least direct evidence: we report no controlled measurement of integration effort, and \S\ref{sec:future} treats that as open.

\subsection{Training smoke tests}
\label{sec:smoke}

Table~\ref{tab:demos} gives three training smoke tests on a single workstation GPU. Each fine-tunes a small model to adopt an output contract it ignores at baseline, so a before/after generation shows the weights moved. These exercise the trainer contract end to end; they are not competitive benchmark results.

\begin{table}[tb]
\centering
\small
\setlength{\tabcolsep}{4pt}
\begin{tabular}{p{2.05cm}p{2.35cm}p{2.6cm}}
\toprule
Path & Setup & Result \\
\midrule
Distillation (SFT) & SmolLM2-360M \citep{allal2025smollm2}, 16 teacher trajectories, LoRA, $\sim$7\,s & train loss $3.60{\to}0.25$; token acc.\ $0.47{\to}0.94$; contract learned \\
\addlinespace
GRPO (on-policy) & SmolLM2-360M, 90 steps, contract-only reward & mean reward $0.03{\to}0.66$; contract rate $25\%{\to}100\%$; correct-label $0\%{\to}75\%$ \\
\addlinespace
Self-heal $\rightarrow$ DPO & 12 corrective preference pairs from the heal loop & reward acc.\ $0.50{\to}1.00$; final reward margin $1.97$ \\
\bottomrule
\end{tabular}
\caption{Representative training runs shipped with \agenttune{}, captured on a single NVIDIA RTX PRO 6000 Blackwell Max-Q workstation GPU. These are smoke tests and the denominators are small: the contract and correct-label rates are four sampled completions on \emph{one} probe input ($1/4 \rightarrow 4/4$ and $0/4 \rightarrow 3/4$), and the distillation row reports \emph{in-sample training loss} at epochs $1.25$ and $25.0$ over the same 16 trajectories, not a held-out result. They demonstrate that each trainer path executes end to end and moves the policy, not that \agenttune{} sets a benchmark.}
\label{tab:demos}
\end{table}

\subsection{The DECIDE closed loop end to end}
\label{sec:loopcs}

The closed loop is verified end to end against a real \code{bfsi/kyc\_triage} audit log that loops. The detector reports six \code{loop\_collapse} failures, and the classifier and generator buffer training examples from them. A self-heal DPO retrain is then gated on accuracy, which rises from $0.33$ to $1.00$ over six held-out cases reconstructed from past successful runs, before a LoRA adapter is deployed and reloaded as a \code{PeftModel}. The gate's trajectory arm was not exercised in this run, so the reported decision rests on task accuracy alone.

\subsection{GPU-free testability and dependency footprint}
\label{sec:repro}

The orchestration surfaces are independently testable through injected models, rollout engines, and runners. Thus \code{Project} lifecycle plumbing, \eventlog{} projections, harness conformance, DECIDE graph compilation, closed-loop triggers, synthesis stages, and deterministic trajectory metrics can be exercised without loading a trainable model; actual RL training and model-backed rollout paths require their corresponding model/GPU or API dependencies. The repository also ships captured training scripts and executable notebooks for the principal paths, with limitations and dependency assumptions recorded alongside the runs.

\section{Use Case}
\label{sec:usecase}

\S\ref{sec:casestudies} showed that each path runs. This section works one domain end to end through all of them at once: a legal contract-search agent carried from a raw corpus, through synthesized training data and a trained search policy, to a distilled student small enough to run on-premises. It is the closest thing in this paper to the workflow the library is built for.

\subsection{Compact legal search agents}
\label{sec:legal}

\begin{figure}[tb]
\centering
\IfFileExists{figures/retrieval-gap.pdf}{\includegraphics[width=0.86\columnwidth]{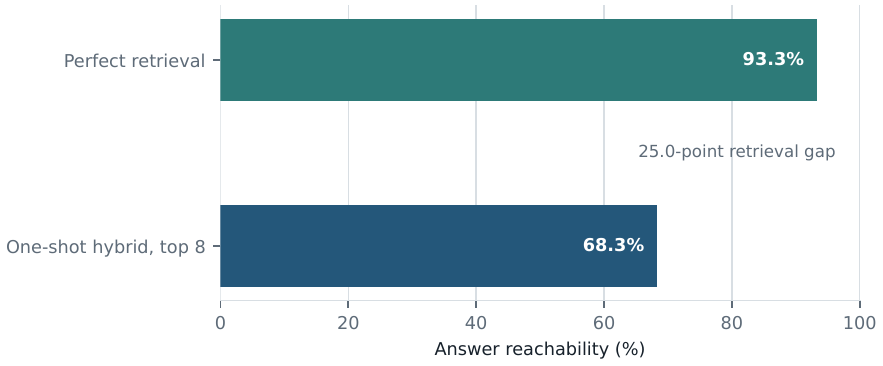}}{\fbox{\parbox[c][0.18\textheight][c]{0.82\columnwidth}{\centering Retrieval-gap figure\\(figures/retrieval-gap.pdf not supplied)}}}
\caption{Answer reachability under one-shot top-8 hybrid retrieval and perfect retrieval on the multi-hop CUAD questions.}
\label{fig:retgap}
\end{figure}

Contract review is a search problem. A reviewer may need to combine a definition in one section with an obligation elsewhere that refers back to it as ``such rate'' or ``the foregoing.'' On multi-hop questions from the CUAD contract corpus \citep{hendrycks2021cuad}, one-shot top-8 hybrid retrieval makes the answer reachable for 68.3\% of questions, compared with 93.3\% under perfect retrieval (Figure~\ref{fig:retgap}). Retrieval therefore makes the answer unavailable for nearly one-third of the questions before generation begins.

We built an end-to-end system for this task with \agenttune{}: the synthesis pipeline (\S\ref{sec:rag}) generated the training data, the agentic-RAG path trained the search policy, and agentic distillation (\S\ref{sec:distill}) compressed it.

\subsubsection{Task and dataset curation}
\begin{figure}[tb]
    \centering
    \IfFileExists{figures/Verified_Multi-Hop_Dataset_Funnel.png}{
        \includegraphics[width=0.9\columnwidth]{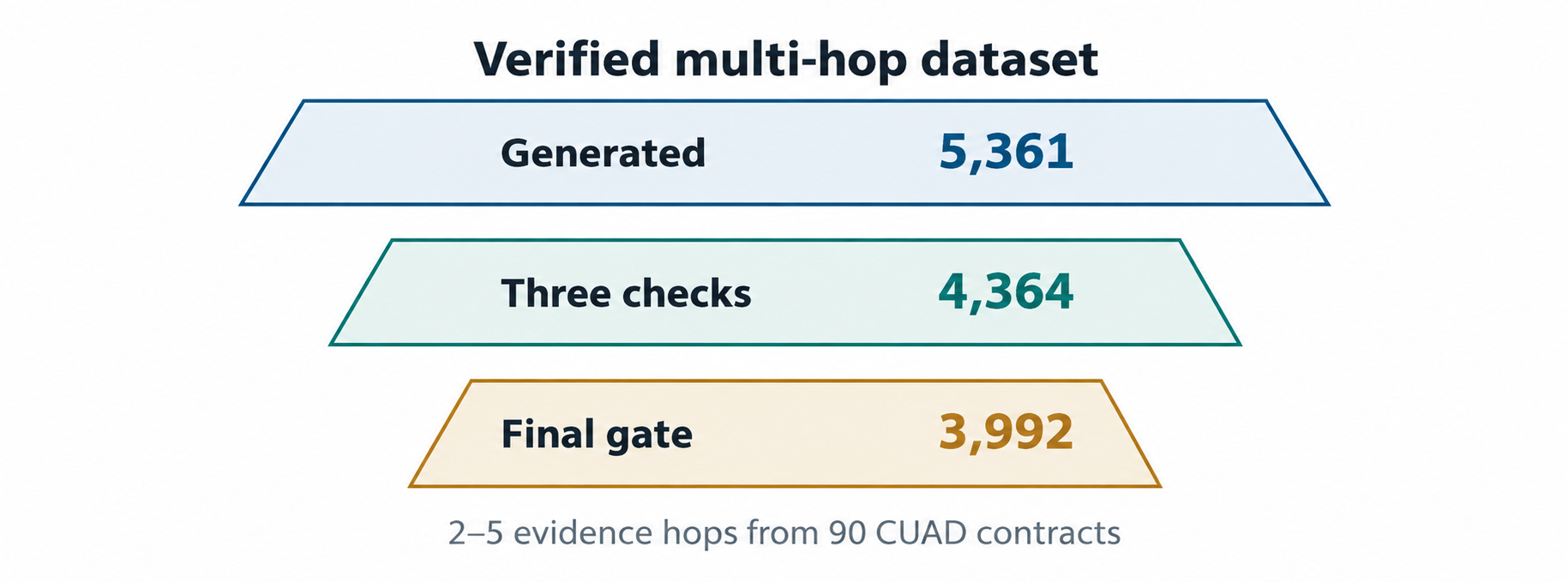}
    }{
        \fbox{
            \parbox[c][0.18\textheight][c]{0.86\columnwidth}{
                \centering
                Synthesis-funnel figure\\
                (figure not supplied)
            }
        }
    }
    \caption{Dataset curation from 5{,}361 candidates to 3{,}992 verified multi-hop questions.}
    \label{fig:funnel}
\end{figure}

From 90 CUAD training contracts, the pipeline generated 5{,}361 candidate questions. Answerability, chain-dependency, and retrieval-necessity verification retained 4{,}364 candidates; a final answerability gate removed 372 more, leaving 3{,}992 questions spanning two to five hops (Figure~\ref{fig:funnel}). The train/validation split is gold-chunk disjoint: questions that share a gold passage remain on the same side, so no evidence passage occurs in both partitions.

\subsubsection{Experimental setup}
\label{sec:legalsetup}
A Qwen3.5-9B \citep{qwen3.5} model is trained with GRPO and LoRA using rank 16, $\alpha=32$, and a learning rate of $10^{-6}$. The agent searches the full 90-contract corpus over several turns and receives rewards for answer correctness, evidence coverage, and search frugality, following the efficiency-oriented design of FrugalRAG \citep{frugalrag} and IKEA \citep{ikea}. Evaluation uses greedy decoding, and the frozen policy is additionally carried, without further training, to three previously unseen LegalBench-RAG corpora \citep{pipitone2024legalbenchrag}. Appendix~\ref{app:legalcfg} collects the full configuration. 

We evaluate answer quality using exact match (EM), token-level F1, precision, recall, and Pass@1. EM requires the predicted answer to match the reference answer exactly, while token-level F1 measures partial overlap between them. Precision measures the correctness of the returned predictions, recall measures their coverage of the reference answers, and Pass@1 records the proportion of examples answered correctly by the first generated response without additional sampling or retries.

Retrieval performance is measured at several levels. R@\(k\) indicates whether the exact gold chunk appears among the top-\(k\) retrieved chunks, while ans@\(k\) indicates whether at least one gold evidence passage is retrieved. Doc@8 measures the fraction of questions for which the gold source document appears among the retrieved passages, and Clause@8 measures whether the relevant gold clause is present in the top-eight results. Answer reachability measures whether the retrieved evidence is sufficient to derive the reference answer and is therefore a retrieval ceiling rather than an end-to-end accuracy score. We additionally report searches per question (S/Q) to measure retrieval cost and search-policy efficiency.

\subsubsection{Comparative analysis}
\label{sec:legalresults}
\begin{figure}[tb]
\centering
\IfFileExists{figures/legal-results.pdf}{\includegraphics[width=\columnwidth]{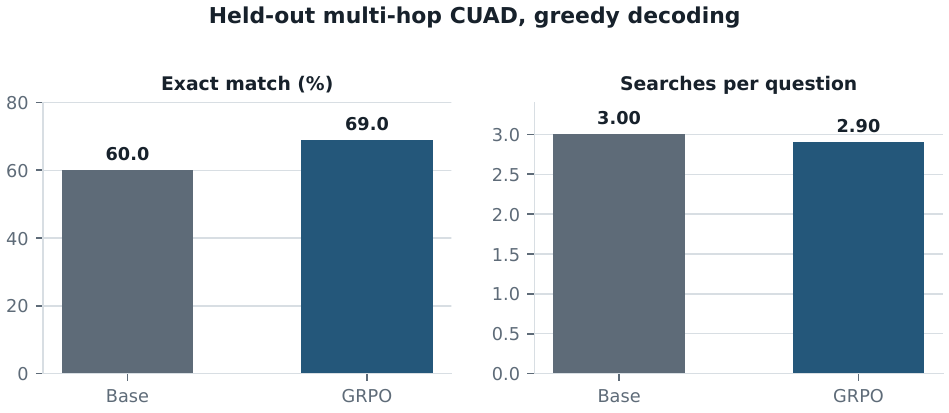}}{\fbox{\parbox[c][0.18\textheight][c]{0.96\columnwidth}{\centering Legal-results figure\\(figures/legal-results.pdf not supplied)}}}
\caption{Greedy-decoding performance on held-out multi-hop CUAD questions. GRPO improves exact match while slightly reducing search use.}
\label{fig:legalres}
\end{figure}

\emph{In-domain.} On held-out multi-hop CUAD questions, exact match rises from 60.0\% for the base model to 69.0\% after GRPO, while searches per question fall from 3.00 to 2.90 (Table~\ref{tab:legalres}, Figure~\ref{fig:legalres}). The two move in the same direction: the trained policy answers more questions correctly while issuing fewer retrievals, which is the frugality effect FrugalRAG \citep{frugalrag} reports.

\begin{figure}[tb]
\centering
\IfFileExists{figures/transfer-results.pdf}{\includegraphics[width=\columnwidth]{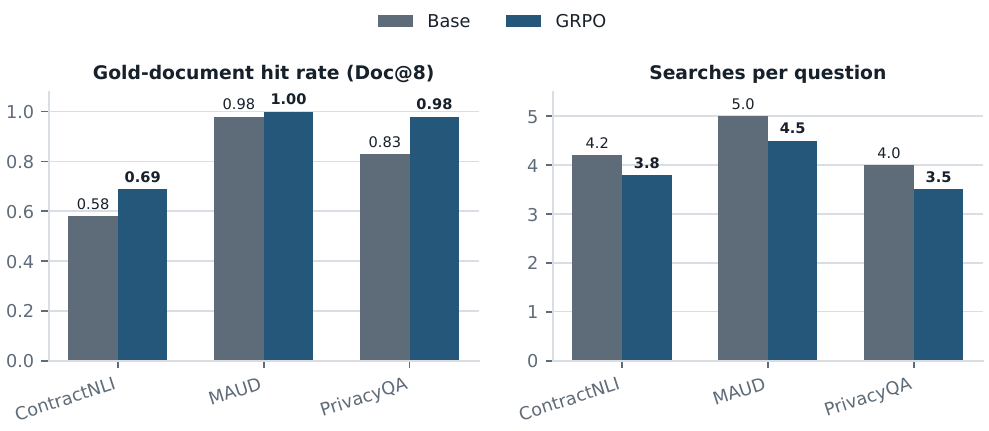}}{\fbox{\parbox[c][0.18\textheight][c]{0.96\columnwidth}{\centering Transfer-results figure\\(figures/transfer-results.pdf not supplied)}}}
\caption{Transfer to three unseen LegalBench-RAG corpora. The frozen GRPO policy improves Doc@8 and reduces searches on each dataset.}
\label{fig:transfer}
\end{figure}

\emph{Transfer.} Table~\ref{tab:transfer} reports the frozen policy on the three unseen corpora. Doc@8 improves and searches fall on every one. On MAUD \citep{wang-etal-2023-maud} the base policy already retrieves the gold document for 98\% of queries, so the move from 0.98 to 1.00 is a ceiling effect and the transferable signal there is the drop in searches from 5.0 to 4.5 rather than the gain in Doc@8. PrivacyQA \citep{ravichander-etal-2019-question} has the most headroom of the three and gains the most, from 0.83 to 0.98. ContractNLI \citep{koreeda-manning-2021-contractnli} sits between them. The pattern is consistent across all three: the search behavior learned on CUAD, not a corpus-specific ranking, is what carries (Figure~\ref{fig:transfer}).

\begin{table}[tb]
\centering
\small
\begin{tabular}{lcc}
\toprule
Model & Exact match (\%) & Searches / question \\
\midrule
Base model              & 60.0 & 3.00 \\
\textbf{+ GRPO (LoRA)}  & \textbf{69.0} & \textbf{2.90} \\
\bottomrule
\end{tabular}
\caption{Held-out multi-hop CUAD questions under greedy decoding. GRPO raises exact match while reducing search use.}
\label{tab:legalres}
\end{table}

\begin{table}[tb]
\centering
\small
\setlength{\tabcolsep}{5pt}
\begin{tabular}{lcccc}
\toprule
& \multicolumn{2}{c}{Doc@8} & \multicolumn{2}{c}{Searches / question} \\
\cmidrule(lr){2-3}\cmidrule(lr){4-5}
Corpus & Base & + GRPO & Base & + GRPO \\
\midrule
ContractNLI \citep{koreeda-manning-2021-contractnli} & 0.58 & \textbf{0.69} & 4.2 & \textbf{3.8} \\
MAUD \citep{wang-etal-2023-maud}                     & 0.98 & \textbf{1.00} & 5.0 & \textbf{4.5} \\
PrivacyQA \citep{ravichander-etal-2019-question}     & 0.83 & \textbf{0.98} & 4.0 & \textbf{3.5} \\
\bottomrule
\end{tabular}
\caption{Transfer of the frozen CUAD-trained policy to three unseen LegalBench-RAG corpora \citep{pipitone2024legalbenchrag}. Doc@8 improves and searches fall on every corpus; the MAUD gain is a ceiling effect.}
\label{tab:transfer}
\end{table}

\subsubsection{Agentic distillation}
The trained 9B policy serves as the teacher for Qwen2.5-0.5B and Qwen2.5-1.5B students. Supervised fine-tuning uses complete teacher trajectories but masks retrieved tool outputs from the loss. The students are therefore trained on the sequential decisions that decompose a question, retrieve an initial fact, refine the query, retrieve the remaining evidence, and combine the result, while reducing the incentive to reproduce retrieved passages. The 1.5B student can run on-premises without external inference calls, and the closed loop can use later failures as corrective training data.

We evaluate trajectory distillation using Qwen2.5 students on two legal benchmarks. The Qwen2.5-0.5B student is evaluated on CUAD, while the Qwen2.5-1.5B student is evaluated on the ECTHR task from LexGLUE. We compare the pretrained student, answer-only supervised fine-tuning, question-answer supervision with generated reasoning, unmasked teacher trajectories, and our masked-trajectory objective.

Table~\ref{tab:distillation-results} shows that the masked objective produces the highest Pass@1 on both benchmarks. On CUAD, masked trajectory training reaches 45.8 Pass@1, improving by 4.8 points over unmasked trajectory supervision and by 9.3 points over question-answer supervision with reasoning. The unmasked variant obtains higher recall, 58.2 compared with 52.6, but substantially lower precision, 39.4 compared with 48.5. This difference indicates that reproducing every part of the trajectory encourages broader but less selective predictions.

On LexGLUE ECTHR, the masked 1.5B student achieves the best result on all three reported metrics: 53.2 precision, 58.8 recall, and 55.4 Pass@1. Its Pass@1 exceeds unmasked trajectory training by 5.9 points and question-answer supervision with reasoning by 11.2 points. Across both evaluations, complete teacher trajectories are more effective when retrieved environment responses remain available as context but do not contribute directly to the training loss. The resulting objective supervises the student's reasoning, tool calls, and final answer without requiring it to reproduce retrieved contract text.

\begin{table}[t]
\centering
\footnotesize
\setlength{\tabcolsep}{4pt}
\renewcommand{\arraystretch}{1.08}
\begin{tabular}{llrrr}
\toprule
Model & Training variant & Precision & Recall & Pass@1 \\
\midrule
\multicolumn{5}{l}{\emph{CUAD}} \\
\addlinespace[2pt]
0.5B & Base                    & 36.8 & 34.1 & 29.4 \\
0.5B & Answer SFT              & 38.2 & 24.1 & 28.5 \\
0.5B & QA + reasoning          & 45.1 & 42.0 & 36.5 \\
0.5B & Trajectory, no mask     & 39.4 & \textbf{58.2} & 41.0 \\
0.5B & Masked trajectory       & \textbf{48.5} & 52.6 & \textbf{45.8} \\
\midrule
\multicolumn{5}{l}{\emph{LexGLUE ECTHR}} \\
\addlinespace[2pt]
1.5B & Base                    & 37.7 & 40.2 & 28.3 \\
1.5B & Answer SFT              & 42.0 & 31.2 & 33.1 \\
1.5B & QA + reasoning          & 39.4 & 48.9 & 44.2 \\
1.5B & Trajectory, no mask     & 46.8 & 54.1 & 49.5 \\
1.5B & Masked trajectory       & \textbf{53.2} & \textbf{58.8} & \textbf{55.4} \\
\bottomrule
\end{tabular}
\caption{Agentic-distillation ablation on CUAD and LexGLUE ECTHR.
The 0.5B student is evaluated on CUAD and the 1.5B student on ECTHR.
Masked trajectory training excludes retrieved tool outputs from the loss
while retaining them as conditioning context. Bold values indicate the
best result within each benchmark.}
\label{tab:distillation-results}
\end{table}

\subsection{Beyond legal: a general pattern}
\label{sec:applications}

Contract search is one instance of a broader class: agents that search across information, decide what to retrieve next, learn from failures, and run in constrained production environments. The same pipeline, synthesis, agentic-RAG training, distillation, and the self-healing loop, applies to workflows in other regulated domains.

\paragraph{BFSI.}
Fraud-detection agents that search transaction histories, route high-risk cases to human review, and learn from false positives. Trade-surveillance agents that combine evidence across systems and adapt as transaction patterns shift.

\paragraph{Healthcare.}
Prior-authorization workflows that retrieve across patient records, combine evidence from multiple visits, and flag missing documentation. Clinical decision support that retrieves relevant guidelines and folds physician feedback into later retraining.

\paragraph{Support.}
Support agents that search documentation, refine queries when the first results fall short, and learn from misrouted or unresolved tickets. Knowledge-base agents that learn which search paths lead to resolution.

The common shape: a small on-premises model, a corpus it searches over several turns, a reward that scores grounded answers and frugal retrieval, and a closed loop that turns production failures into corrective training data. The repository ships scripted spine case studies for each of these domains, run with open models and open data.

\paragraph{Insurance.}
Insurance workflows provide another instance of the same pattern. Fraud investigation and claim-triage agents can search policy and claim records, combine evidence across sources, and route uncertain cases for review; claim-propensity and severity workflows can similarly combine structured prediction with tool-mediated analysis.

\section{Discussion and Related Work}
\label{sec:discussion}

\S\ref{sec:related} places \agenttune{} against the systems it builds on and the lines of work it borrows from, \S\ref{sec:featcmp} states where it differs from the tools it composes with, and \S\ref{sec:design} gives the four decisions that shaped the design.

\subsection{Related work}
\label{sec:related}

\paragraph{RL training libraries.}
TRL \citep{vonwerra2020trl} provides reference trainers for PPO \citep{schulman2017ppo}, DPO \citep{rafailov2023dpo}, GRPO \citep{shao2024deepseekmath}, RLOO \citep{ahmadian2024rloo}, and BCO \citep{jung2024bco}. verl \citep{sheng2024hybridflow} and OpenRLHF \citep{hu2024openrlhf} scale RLHF to large clusters with hybrid-parallel execution engines. These libraries optimize a policy over completions. Agentic tool use, where one training example is a multi-turn transcript of model text and tool outputs, falls to the caller. \agenttune{} builds on TRL, dispatches to its trainers through one factory, and adds the multi-turn rollout loop, the reward composition layer, and the trajectory schema around them. It does not reimplement the optimizers, and it inherits TRL's single-node scaling limits.

\paragraph{Agent frameworks and agentic RL.}
ReAct \citep{yao2023react}, Reflexion \citep{shinn2023reflexion}, Plan-and-Solve \citep{wang2023planandsolve}, and Tree-of-Thoughts \citep{yao2023tot} are control policies for the reason--act loop; LangGraph \citep{langgraph2024} composes such loops into stateful graphs; Toolformer \citep{schick2023toolformer} and the ReAct line established tool-augmented generation, and \citet{yang2023autogptonlinedecisionmaking} benchmarks the autonomous-agent form of the same loop. Search-R1 \citep{jin2025searchr1} trains a policy to interleave reasoning with retrieval and masks retrieved tokens from the loss. \agenttune{} treats the strategies as a swappable layer behind one interface, adopts retrieved-token masking as a schema-level property of the \eventlog{}, and makes any strategy trainable through the shared core.

Closest in name is AgentTuning \citep{Zeng2023AgentTuningEG}, which fine-tunes LLMs on agent interaction trajectories and releases both a trajectory dataset and the resulting models. The contributions are complementary rather than competing: their unit of delivery is data and weights for generalized agent ability, ours is a schema and a lifecycle that a team applies to its own agent. AgentTuning answers what to train on; \agenttune{} answers how to keep a trajectory usable from collection through recovery.

\paragraph{Distillation and synthetic data.}
Knowledge distillation \citep{hinton2015distilling} and its task-specific \citep{sanh2020distilbertdistilledversionbert} and sequence-level \citep{gu2024minillm} variants compress models; Self-Instruct \citep{wang2023selfinstruct} and successors \citep{honovich-etal-2023-unnatural} generate instruction data. Agent distillation typically trains on static input--output pairs. \agenttune{} distills the \emph{trajectory}, the ordered decisions of when to call a tool and when to stop, with environment responses masked out of the loss, so the student learns the teacher's procedure and not its retrieved text.

\paragraph{Agent evaluation and RAG synthesis.}
The \code{lm-eval} harness \citep{eval-harness} is the standard for static benchmarks. RAGAS \citep{es2024ragas} scores RAG faithfulness; GRADE \citep{grade2024} calibrates multi-hop retrieval difficulty from a typed passage graph; HotpotQA \citep{yang2018hotpotqa} and MuSiQue \citep{trivedi} established the multi-hop QA setting, and IRCoT \citep{trivedi2023interleaving} interleaves retrieval with chain-of-thought reasoning. \agenttune{} wraps the \code{lm-eval} CLI for static tasks, adds deterministic rule-based trajectory metrics (tool-argument correctness, redundancy, loop-collapse, plan adherence) for procedural quality, and reuses the RAGAS/GRADE primitives in its synthesis pipeline.

\paragraph{Agentic search and legal retrieval.}
The search-agent line is the closest published analogue to \S\ref{sec:legal}. Search-R1 \citep{jin2025searchr1}, DeepRetrieval \citep{jiang2025deepretrieval} and FrugalRAG \citep{frugalrag} train retrieval policies with RL, the latter arguing that fewer, better-targeted searches beat exhaustive ones --- the effect Figure~\ref{fig:legalres} reproduces. IKEA \citep{ikea} learns when to consult an external index at all rather than relying on parametric knowledge, and \citet{liu2026searchdonotguess} makes abstention on insufficient evidence an explicit training objective. On the domain side, LegalSearch-R1 \citep{fan2026legalsearchr1} applies RL to legal retrieval directly, query rewriting for specialist corpora is studied by \citet{li2026thinkthenrewrite}, LegalBench \citep{guha2023legalbenchcollaborativelybuiltbenchmark} and LexGLUE \citep{chalkidis2022lexgluebenchmarkdatasetlegal} supply the task inventory, \citet{qiu2025legalqa} pairs LLM-generated data with distillation for legal question answering much as \S\ref{sec:legal} does, \citet{hou2025largelanguagemodelsmeet} surveys the wider area, and multi-agent legal assistants such as ChatLaw \citep{Cui_2026} target the same practitioner workflow through orchestration rather than training. \agenttune{} contributes no new retrieval algorithm to this line; it supplies the substrate the line needs, namely one trajectory record that a search policy can be trained on, evaluated from, and repaired through.

\paragraph{Decision workflows and self-healing.}
Teams increasingly express production LLM decisioning as declarative pipelines with audit trails. \agenttune{}'s DECIDE engine compiles a YAML workflow into a LangGraph \citep{langgraph2024} state graph, and its closed loop wires that audit trail back to the RL core. A classifier and a generator turn failures into preference data, a trigger decides when enough has accumulated, and a deployment gate scores the old and new models on a held-out set rebuilt from past successful runs before anything ships. It resembles continuous-training MLOps, specialized to agent trajectories: the unit of feedback is a multi-turn transcript with a classified root cause, and the gate checks procedural regressions (new loops, worse tool-argument accuracy) alongside end-task accuracy.

\subsection{Feature comparison}
\label{sec:featcmp}
Table~\ref{tab:related} places \agenttune{} against the tools it composes with. The comparison is deliberately not a ranking: \agenttune{} occupies a different axis from the columns it sits beside. TRL and verl compete on optimizer throughput and scale, LangGraph on execution semantics, \code{lm-eval} on benchmark coverage, and \agenttune{} builds on the first and the third rather than replacing them. Its axis is lifecycle continuity --- it is the one row where a single trajectory schema spans every stage, so an artifact moves from orchestration to RL to distillation to monitoring without a per-boundary adapter. A team that needs the fastest optimizer should reach for the columns; a team that needs the same trajectory to survive five stages is the case this row is for.

\begin{table}[tb]
\centering
\scriptsize
\setlength{\tabcolsep}{3.5pt}
\renewcommand{\arraystretch}{1.15}
\begin{tabular}{p{3.4cm}ccccc}
\toprule
Capability & \rotatebox{90}{\agenttune{}} & \rotatebox{90}{TRL} & \rotatebox{90}{verl} & \rotatebox{90}{SkyRL / OpenRLHF} & \rotatebox{90}{LangChain / CrewAI} \\
\midrule
Trains a model at all                          & \checkmark & \checkmark & \checkmark & \checkmark & -- \\
GRPO/PPO/DPO/RLOO/BCO via one entry point       & \checkmark & --         & partial    & partial    & -- \\
Agentic tool-calling rollouts native to every algorithm & \checkmark & --  & \checkmark & \checkmark & executes tools, no training \\
One trajectory schema across eval / train / distill / heal & \checkmark & -- & -- & -- & -- \\
Agentic distillation (behavior, not weights alone) & \checkmark & -- & -- & -- & -- \\
Synthetic multi-hop data with verification       & \checkmark & -- & -- & -- & -- \\
Decision workflows as YAML                        & \checkmark & -- & -- & -- & code, not config \\
Self-healing, gated production retraining         & \checkmark & -- & -- & -- & -- \\
\midrule
License                                           & LSAL 1.1 & Apache 2.0 & Apache 2.0 & Apache 2.0 & MIT \\
Adapters to connect $n$ lifecycle stages          & 6 projections & \multicolumn{4}{c}{up to $n(n-1)/2 = 15$ pairwise converters} \\
\bottomrule
\end{tabular}
\caption{\agenttune{} against the tools it composes with, assessed in September 2026 against each project's then-current documented feature set. TRL supplies the optimizers; verl, SkyRL \citep{skyrl2025}, and OpenRLHF scale RL to clusters; LangChain \citep{langchain2022} and CrewAI \citep{crewai2024} orchestrate agents without a training loop. \agenttune{} reuses TRL and LangGraph rather than reimplementing them, and adds the schema and the stages built on it. Entries record whether a capability is exposed as a first-class, documented entry point, not whether it can be assembled with user code; ``partial'' means some but not all of the five algorithms. The last row is the one quantitative cell: connecting six lifecycle stages pairwise would need up to fifteen format converters, where one schema needs six projections. \agenttune{} is source-available under LSAL~1.1 and is therefore \emph{not} open source in the OSI sense, unlike every other column.}
\label{tab:related}
\end{table}

\subsection{Design decisions}
\label{sec:design}

\paragraph{The schema is the only fixed contract.}
A monolithic framework that owns the strategy, the tools, the trainer, and the monitor forces a rewrite when a user wants a different orchestration library or a different trainer. \agenttune{} couples on the \eventlog{}. A strategy is any object with four methods; a tool is any object with \code{.name} and \code{.to\_schema()}, or a plain annotated callable; a trainer is any callable returning something with a \code{.train()} method; a judge is any object with an \code{evaluate\_trajectory} method. The library ships defaults for each.

\paragraph{One place for the loss mask.}
Segment-aware, completion-only loss masking is usually implemented inside a trainer. The SFT projection generates the mask once from the same event kinds read by evaluation and failure detection. Project training, distillation, closed-loop retraining, and custom SFT code therefore share one definition of which tokens belong to the model.

\paragraph{Injected boundaries.}
The trainer, the rollout engine, the LLM judge, the retrain job, and the deployment-gate scorer are parameters, not imports. The closed-loop orchestration, the distillation pipeline, and the multi-agent graph run their full control flow under unit tests with fake callables, and a user swaps a hosted judge for a local one without touching the orchestration. Building an episode, projecting an \eventlog{}, previewing a distillation dataset, compiling a DECIDE graph, and running the synthesis pipeline with fake clients all work with no GPU and no model.

\paragraph{Captured runs and a known-issues list.}
Every real-model example in the repository ships its captured output and its caveats, and gaps found while testing go on a \emph{Known Issues} page. \S\ref{sec:limitations} draws from that page.

\section{Limitations}
\label{sec:limitations}

\paragraph{Scale.}
\agenttune{} builds on TRL and inherits its single-node scaling limits. It is not a cluster-scale RLHF system like verl or OpenRLHF. The shipped demonstrations are small contract-learning smoke tests: they make a weight change visible and cheap to reproduce, but they are not competitive results on public leaderboards.

\paragraph{Training paths.}
The standalone SFT path under \code{core/sft} is not wired end to end; agentic RL and the \code{trainer\_factory}-based distillation path are the training paths that run, and every worked example wraps TRL's trainers directly. The CLI has no \code{-{}-tools} flag, so the library's headline capability, tool-calling training, is not reachable from the command line.

\paragraph{Tooling and packaging.}
Fetching \emph{any} builtin tool through \code{ToolRegistry.get(\dots)} triggers a blanket import of every builtin module, so in a lean environment without \code{langchain\_community} even the pure-stdlib tools fail; importing a tool class directly sidesteps this. Some builtin tools (GitHub, Slack, Playwright) need credentials or browser binaries and have no test coverage in the repository. \code{RougeMetric}/\code{BleuMetric} return silent zeros if \code{rouge\_score}/\code{nltk} are absent.

\paragraph{Reward and collection defects.}
One default reward configuration currently scores a well-formed, a malformed, and a missing tool call identically, and the FinQA/use-case reward modules have a name-shadowing subtlety, so the stronger scorers are reachable only under prefixed keys. Three further gaps belong here rather than only on the \emph{Known Issues} page: \code{StageWiseRunner} silently ignores conditional routing, so it is not a substitute for \code{GraphRunner} on any branching template; \code{AuditReader.extract\_dpo\_pairs} and \code{DecideToTrainerBridge.extract\_bco\_labels} return empty against an audit log, which is why \code{CollectRunner} --- reading each episode's in-memory \code{PipelineState} --- is the working collection path; and \code{distilled\_judge\_reward}, the in-loop scorer \S\ref{sec:rlcore} recommends, returns zeros on any exception rather than raising, contradicting the principle stated for the reward layer. The \emph{Known Issues} page documents each of these with its code-level cause.

\paragraph{Domain validation.}
The RAG synthesis pipeline and the trajectory metrics are validated on English document corpora; multilingual and non-contractual domains are future work.

\paragraph{Evaluation.}
Finally, a limitation of the evaluation rather than of the library: \emph{every number reported in this paper is an internal comparison}. Table~\ref{tab:demos} contrasts \agenttune{}-trained weights against their own base checkpoints, \S\ref{sec:legal} does the same on a question set this library synthesized, and Table~\ref{tab:related} compares capabilities without measuring them. No result here is scored by a harness we did not write, because the release contains no agentic-benchmark integration --- neither the tool-agent-user benchmarks ($\tau$-bench \citep{taubench2024}, ToolBench \citep{toolllm2023}) nor the broader agentic suites (AgentBench \citep{agentbench2023}, WebArena \citep{webarena2023}) are wired to the harness. A single public calibration point on any of them is the most useful next measurement, and \S\ref{sec:future} treats it as such.

\section{Conclusion}
\label{sec:conclusion}

\agenttune{} is designed to support the agent development lifecycle through one trajectory schema rather than separate representations for execution, training, evaluation, and recovery. The \eventlog{} projects in from RL rollouts, DECIDE runs, and evaluation traces and out to training rows, evaluator inputs, and audit records. Around this contract, the library provides an agentic runtime, a TRL-backed RL core with multi-turn tool-calling rollouts, agentic distillation and RAG synthesis, YAML decision workflows, and a designed self-healing path. The reported experiments establish representative components separately; the legal case study provides the strongest task-level evidence.

The library is available at \url{https://github.com/Lexsi-Labs/AgentTune}, released under the Lexsi Labs Source Available License (LSAL) v1.1.

\section{Future Work}
\label{sec:future}

Several directions are open. Multilingual and non-contractual domains need verification, and naturally occurring user queries remain to be tested. Tool isolation runs at process level today; container and remote-host isolation are planned. The standalone \code{core/sft} path is not wired end to end, and a generated API reference would replace the current hand-maintained one. Cluster-scale RL, which \agenttune{} does not attempt, is left to verl and OpenRLHF. The most valuable single addition is external calibration: one public tool-use benchmark scored through \code{create\_agentic\_trainer}, which would place this paper's internal numbers on a scale a reader already knows.

\section{Security and Ethical Considerations}
\label{sec:ethics}

\paragraph{Threat model.} \agenttune{} trains agents that execute code and call external services, and the shipped defaults favor usability over confinement. Two builtin tools are unrestricted by construction: \code{RunBashTool} passes a model-generated string to \code{subprocess.run(\dots, shell=True)}, and \code{RunPythonTool} executes model-generated source with the full interpreter. Both inherit the credentials, network access and filesystem of the training process. \code{DictToolHarness} runs callables in-process with no sandbox at all, and its snapshot/restore path uses in-process pickle blobs, so a harness snapshot is executable data and must not cross a trust boundary. \code{IsolatedTool} raises the floor to a separate OS process, but that is Level~1 of a planned ladder --- process isolation, not a container or a VM --- and it falls back to in-process execution silently unless the caller passes \code{require\_isolation=True}, which defaults to \code{False}. The practical guidance is that untrusted tools and untrusted generated code belong behind \code{OpenEnvHarness}, whose sandbox is remote, and that a deployment exposing \code{RunBashTool} to an untrusted prompt should be treated as offering remote code execution. \code{SECURITY.md} in the repository carries the disclosure process and the current per-component status.

\paragraph{Injection and feedback-loop integrity.} Two risks follow from the closed loop rather than from any single tool. A retrieved document is untrusted input that reaches the policy verbatim, so indirect prompt injection through a corpus is a live concern for the search agents of \S\ref{sec:legal}; the trajectory metrics detect anomalous tool use but do not detect instructions embedded in retrieved text. More seriously, the self-healing loop closes over its own outputs: failures are classified, turned into preference pairs, and fed back as training data. An adversary who can influence production inputs can therefore influence the training buffer, and the deployment gate is an accuracy check rather than a provenance check --- it will approve a model that improved on a poisoned held-out set. Deployments that accept untrusted traffic should treat the retraining buffer as attacker-influenced, and should curate it before it reaches \code{DecideToTrainerBridge}.

\paragraph{Data protection.} Audit records are written verbatim and are not redacted. The BFSI KYC example in Listing~\ref{lst:collect} carries name, date of birth, employer and income through the audit trail and into the retraining buffer, and \code{AuditWriter} exposes no redaction hook, so any pipeline over personal data needs a redaction stage in front of it --- a role the PII detector of CuratorKIT \citep{curatorkit2026} is designed to fill. Synthetic corpora and LLM judges inherit the biases of the models that produced them, and both the synthesis pipeline and the trajectory metrics here are validated only on English contractual text; neither the difficulty calibration nor the judge rubric should be assumed to transfer to another language or domain without revalidation.

\paragraph{Intended use.} The shipped domain templates are demonstrations of the workflow engine, not certified decision systems. The BFSI, healthcare and legal workflows encode plausible rather than regulator-approved logic, and using one to make consequential decisions about a person would require domain validation, an approval process and a human accountable for the outcome --- none of which the library supplies. Two properties help and should not be overstated: \code{human\_review} fails closed, so a review that times out is denied, not approved, and redeployment is gated on measured accuracy. But the gate is automatic, not human-supervised: nothing in the loop requires a person to approve a model before it serves traffic, and an operator who wants that must add it. Regulated deployments should treat the audit trail as the input to human oversight rather than as a substitute for it.

\section*{Acknowledgments}
\agenttune{} builds on HuggingFace Transformers, TRL, HuggingFace Datasets, LangGraph, and OpenEnv.

\bibliographystyle{unsrtnat}
\bibliography{custom}

@article{qiu2025legalqa,
author = {Italiani, Paolo and Moro, Gianluca and Ragazzi, Luca},
year = {2025},
month = {07},
pages = {1-26},
title = {Enhancing legal question answering with data generation and knowledge distillation from large language models},
journal = {Artificial Intelligence and Law},
doi = {10.1007/s10506-025-09463-9}
}

@misc{jin2025searchr1,
      title={Search-R1: Training LLMs to Reason and Leverage Search Engines with Reinforcement Learning}, 
      author={Bowen Jin and Hansi Zeng and Zhenrui Yue and Jinsung Yoon and Sercan Arik and Dong Wang and Hamed Zamani and Jiawei Han},
      year={2025},
      eprint={2503.09516},
      archivePrefix={arXiv},
      primaryClass={cs.CL},
      url={https://arxiv.org/abs/2503.09516}, 
}

@misc{hou2025largelanguagemodelsmeet,
      title={Large Language Models Meet Legal Artificial Intelligence: A Survey}, 
      author={Zhitian Hou and Zihan Ye and Nanli Zeng and Tianyong Hao and Kun Zeng},
      year={2025},
      eprint={2509.09969},
      archivePrefix={arXiv},
      primaryClass={cs.CL},
      url={https://arxiv.org/abs/2509.09969}, 
}

@inproceedings{hendrycks2021cuad,
  title={{CUAD}: An Expert-Annotated {NLP} Dataset for Legal Contract Review},
  author={Hendrycks, Dan and Burns, Collin and Chen, Anya and Ball, Spencer},
  booktitle={Advances in Neural Information Processing Systems (Datasets and Benchmarks Track)},
  volume={34},
  year={2021},
  url={https://datasets-benchmarks-proceedings.neurips.cc/paper/2021/hash/6ea9ab1baa0efb9e19094440c317e21b-Abstract-round1.html}
}

@inproceedings{chalkidis2022lexgluebenchmarkdatasetlegal,
    title = "{L}ex{GLUE}: A Benchmark Dataset for Legal Language Understanding in {E}nglish",
    author = "Chalkidis, Ilias and Jana, Abhik and Hartung, Dirk and Bommarito, Michael and Androutsopoulos, Ion and Katz, Daniel and Aletras, Nikolaos",
    booktitle = "Proceedings of the 60th Annual Meeting of the Association for Computational Linguistics (Volume 1: Long Papers)",
    year = "2022",
    publisher = "Association for Computational Linguistics",
}

@misc{qwen3.5,
    title  = {{Qwen3.5}: Towards Native Multimodal Agents},
    author = {{Qwen Team}},
    month  = {February},
    year   = {2026},
    url    = {https://qwen.ai/blog?id=qwen3.5}
}

@article{yang2024qwen25technicalreport,
  title={Qwen2.5 Technical Report},
  author={{Qwen Team} and An Yang and Baosong Yang and Beichen Zhang and Binyuan Hui and Bo Zheng and Bowen Yu and Chengyuan Li and Dayiheng Liu and Fei Huang and Guanting Dong and Haoran Wei and Huan Lin and Jian Yang and Jianhong Tu and Jianwei Zhang and Jianxin Yang and Jiaxin Yang and Jingren Zhou and Junyang Lin and Kai Dang and Keming Lu and Keqin Bao and Kexin Yang and Le Yu and Mei Li and Mingfeng Xue and Pei Zhang and Qin Zhu and Rui Men and Runji Lin and Tianhao Li and Tingyu Xia and Xingzhang Ren and Xuancheng Ren and Yang Fan and Yang Su and Yichang Zhang and Yunyang Wan and Yuqiong Liu and Zeyu Cui and Zhenru Zhang and Zihan Qiu and Shanghaoran Quan and Zekun Wang},
  journal={arXiv preprint arXiv:2412.15115},
  year={2024}
}

@inproceedings{min,
    title = "Compositional Questions Do Not Necessitate Multi-hop Reasoning",
    author = "Min, Sewon  and
      Wallace, Eric  and
      Singh, Sameer  and
      Gardner, Matt  and
      Hajishirzi, Hannaneh  and
      Zettlemoyer, Luke",
    editor = "Korhonen, Anna  and
      Traum, David  and
      M{\`a}rquez, Llu{\'i}s",
    booktitle = "Proceedings of the 57th Annual Meeting of the Association for Computational Linguistics",
    month = jul,
    year = "2019",
    address = "Florence, Italy",
    publisher = "Association for Computational Linguistics",
    url = "https://aclanthology.org/P19-1416/",
    doi = "10.18653/v1/P19-1416",
    pages = "4249--4257",}

@inproceedings{yang2018hotpotqa,
    title = "{H}otpot{QA}: A Dataset for Diverse, Explainable Multi-hop Question Answering",
    author = "Yang, Zhilin  and
      Qi, Peng  and
      Zhang, Saizheng  and
      Bengio, Yoshua  and
      Cohen, William  and
      Salakhutdinov, Ruslan  and
      Manning, Christopher D.",
    editor = "Riloff, Ellen  and
      Chiang, David  and
      Hockenmaier, Julia  and
      Tsujii, Jun{'}ichi",
    booktitle = "Proceedings of the 2018 Conference on Empirical Methods in Natural Language Processing",
    month = oct # "-" # nov,
    year = "2018",
    address = "Brussels, Belgium",
    publisher = "Association for Computational Linguistics",
    url = "https://aclanthology.org/D18-1259/",
    doi = "10.18653/v1/D18-1259",
    pages = "2369--2380",}

@article{trivedi,
    title = "{M}u{S}i{Q}ue: Multihop Questions via Single-hop Question Composition",
    author = "Trivedi, Harsh  and
      Balasubramanian, Niranjan  and
      Khot, Tushar  and
      Sabharwal, Ashish",
    editor = "Roark, Brian  and
      Nenkova, Ani",
    journal = "Transactions of the Association for Computational Linguistics",
    volume = "10",
    year = "2022",
    address = "Cambridge, MA",
    publisher = "MIT Press",
    url = "https://aclanthology.org/2022.tacl-1.31/",
    doi = "10.1162/tacl_a_00475",
    pages = "539--554",}

@inproceedings{grade2024,
    title = "{GRADE}: Generating multi-hop {QA} and fine-g{RA}ined Difficulty matrix for {RAG} Evaluation",
    author = "Lee, Jeongsoo  and
      Kwon, Daeyong  and
      Jin, Kyohoon",
    editor = "Christodoulopoulos, Christos  and
      Chakraborty, Tanmoy  and
      Rose, Carolyn  and
      Peng, Violet",
    booktitle = "Findings of the Association for Computational Linguistics: EMNLP 2025",
    month = nov,
    year = "2025",
    address = "Suzhou, China",
    publisher = "Association for Computational Linguistics",
    url = "https://aclanthology.org/2025.findings-emnlp.236/",
    doi = "10.18653/v1/2025.findings-emnlp.236",
    pages = "4405--4424",
    ISBN = "979-8-89176-335-7",}

@inproceedings{frugalrag,
title={Frugal{RAG}: Less is More in {RL} Finetuning for Multi-hop Question Answering},
author={Abhinav Java and Srivathsan Koundinyan and Nagarajan Natarajan and Amit Sharma},
booktitle={The Fourteenth International Conference on Learning Representations},
year={2026},
url={https://openreview.net/forum?id=uQKtwdJN0o}
}

@misc{ikea,
      title={Reinforced Internal-External Knowledge Synergistic Reasoning for Efficient Adaptive Search Agent}, 
      author={Ziyang Huang and Xiaowei Yuan and Yiming Ju and Jun Zhao and Kang Liu},
      year={2025},
      eprint={2505.07596},
      archivePrefix={arXiv},
      primaryClass={cs.CL},
      url={https://arxiv.org/abs/2505.07596}, 
}

@inproceedings{lewis2021retrievalaugmentedgenerationknowledgeintensivenlp,
  title={Retrieval-Augmented Generation for Knowledge-Intensive {NLP} Tasks},
  author={Lewis, Patrick and Perez, Ethan and Piktus, Aleksandra and Petroni, Fabio and Karpukhin, Vladimir and Goyal, Naman and K{\"u}ttler, Heinrich and Lewis, Mike and Yih, Wen-tau and Rockt{\"a}schel, Tim and Riedel, Sebastian and Kiela, Douwe},
  booktitle={Advances in Neural Information Processing Systems},
  volume={33},
  year={2020},
  url={https://proceedings.neurips.cc/paper_files/paper/2020/file/6b493230205f780e1bc26945df7481e5-Paper.pdf}
}

@article{ram-etal-2023-context,
  title={In-Context Retrieval-Augmented Language Models},
  author={Ram, Ori and Levine, Yoav and Dalmedigos, Itay and Muhlgay, Dor and Shashua, Amnon and Leyton-Brown, Kevin and Shoham, Yoav},
  journal={Transactions of the Association for Computational Linguistics},
  volume={11},
  pages={1316--1331},
  year={2023},
  doi={10.1162/tacl_a_00605},
  url={https://aclanthology.org/2023.tacl-1.75/}
}

@inproceedings{schick2023toolformer,
 author = {Schick, Timo and Dwivedi-Yu, Jane and Dessi, Roberto and Raileanu, Roberta and Lomeli, Maria and Hambro, Eric and Zettlemoyer, Luke and Cancedda, Nicola and Scialom, Thomas},
 booktitle = {Advances in Neural Information Processing Systems},
 doi = {10.52202/075280-2997},
 editor = {A. Oh and T. Naumann and A. Globerson and K. Saenko and M. Hardt and S. Levine},
 pages = {68539--68551},
 publisher = {Curran Associates, Inc.},
 title = {Toolformer: Language Models Can Teach Themselves to Use Tools},
 url = {https://proceedings.neurips.cc/paper_files/paper/2023/file/d842425e4bf79ba039352da0f658a906-Paper-Conference.pdf},
 volume = {36},
 year = {2023}
}

@misc{yao2023react,
      title={ReAct: Synergizing Reasoning and Acting in Language Models}, 
      author={Shunyu Yao and Jeffrey Zhao and Dian Yu and Nan Du and Izhak Shafran and Karthik Narasimhan and Yuan Cao},
      year={2023},
      eprint={2210.03629},
      archivePrefix={arXiv},
      primaryClass={cs.CL},
      url={https://arxiv.org/abs/2210.03629}, 
}

@article{Cui_2026,
   title={Chatlaw: A Multi-Agent Legal Assistant based on a Role-Aligned Mixture-of-Experts Architecture},
   ISSN={2667-3258},
   url={http://dx.doi.org/10.1016/j.fmre.2026.03.026},
   DOI={10.1016/j.fmre.2026.03.026},
   journal={Fundamental Research},
   publisher={Elsevier BV},
   author={Cui, Jiaxi and Ning, Munan and Li, Zongjian and Li, Hao and Ya, Yang and Chen, Bohua and Ling, Bin and Tian, Yonghong and Yuan, Li},
   year={2026},
   month=May }

@inproceedings{guha2023legalbenchcollaborativelybuiltbenchmark,
  title={LegalBench: A Collaboratively Built Benchmark for Measuring Legal Reasoning in Large Language Models},
  author={Guha, Neel and Nyarko, Julian and Ho, Daniel E. and R{\'e}, Christopher and Chilton, Adam and others},
  booktitle={Advances in Neural Information Processing Systems (Datasets and Benchmarks Track)},
  volume={36},
  year={2023},
  url={https://proceedings.neurips.cc/paper_files/paper/2023/hash/89e44582fd28ddfea1ea4dcb0ebbf4b0-Abstract-Datasets_and_Benchmarks.html}
}

@misc{yang2023autogptonlinedecisionmaking,
      title={Auto-GPT for Online Decision Making: Benchmarks and Additional Opinions}, 
      author={Hui Yang and Sifu Yue and Yunzhong He},
      year={2023},
      eprint={2306.02224},
      archivePrefix={arXiv},
      primaryClass={cs.AI},
      url={https://arxiv.org/abs/2306.02224}, 
}

@misc{hinton2015distilling,
      title={Distilling the Knowledge in a Neural Network}, 
      author={Geoffrey Hinton and Oriol Vinyals and Jeff Dean},
      year={2015},
      eprint={1503.02531},
      archivePrefix={arXiv},
      primaryClass={stat.ML},
      url={https://arxiv.org/abs/1503.02531}, 
}

@misc{sanh2020distilbertdistilledversionbert,
      title={DistilBERT, a distilled version of BERT: smaller, faster, cheaper and lighter}, 
      author={Victor Sanh and Lysandre Debut and Julien Chaumond and Thomas Wolf},
      year={2020},
      eprint={1910.01108},
      archivePrefix={arXiv},
      primaryClass={cs.CL},
      url={https://arxiv.org/abs/1910.01108}, 
}

@inproceedings{wang2023selfinstruct,
    title = "Self-Instruct: Aligning Language Models with Self-Generated Instructions",
    author = "Wang, Yizhong  and
      Kordi, Yeganeh  and
      Mishra, Swaroop  and
      Liu, Alisa  and
      Smith, Noah A.  and
      Khashabi, Daniel  and
      Hajishirzi, Hannaneh",
    editor = "Rogers, Anna  and
      Boyd-Graber, Jordan  and
      Okazaki, Naoaki",
    booktitle = "Proceedings of the 61st Annual Meeting of the Association for Computational Linguistics (Volume 1: Long Papers)",
    month = jul,
    year = "2023",
    address = "Toronto, Canada",
    publisher = "Association for Computational Linguistics",
    url = "https://aclanthology.org/2023.acl-long.754/",
    doi = "10.18653/v1/2023.acl-long.754",
    pages = "13484--13508"
}

@inproceedings{honovich-etal-2023-unnatural,
    title = "Unnatural Instructions: Tuning Language Models with (Almost) No Human Labor",
    author = "Honovich, Or and Scialom, Thomas and Levy, Omer and Schick, Timo",
    booktitle = "Proceedings of the 61st Annual Meeting of the Association for Computational Linguistics (Volume 1: Long Papers)",
    year = "2023",
    address = "Toronto, Canada",
    publisher = "Association for Computational Linguistics",
    url = "https://aclanthology.org/2023.acl-long.806/",
    doi = "10.18653/v1/2023.acl-long.806",
}

@inproceedings{Zeng2023AgentTuningEG,
    title = "{A}gent{T}uning: Enabling Generalized Agent Abilities for {LLM}s",
    author = "Zeng, Aohan  and
      Liu, Mingdao  and
      Lu, Rui  and
      Wang, Bowen  and
      Liu, Xiao  and
      Dong, Yuxiao  and
      Tang, Jie",
    editor = "Ku, Lun-Wei  and
      Martins, Andre  and
      Srikumar, Vivek",
    booktitle = "Findings of the Association for Computational Linguistics: ACL 2024",
    month = aug,
    year = "2024",
    address = "Bangkok, Thailand",
    publisher = "Association for Computational Linguistics",
    url = "https://aclanthology.org/2024.findings-acl.181/",
    doi = "10.18653/v1/2024.findings-acl.181",
    pages = "3053--3077"
}

@inproceedings{trivedi2023interleaving,
    title = "Interleaving Retrieval with Chain-of-Thought Reasoning for Knowledge-Intensive Multi-Step Questions",
    author = "Trivedi, Harsh and Balasubramanian, Niranjan and Khot, Tushar and Sabharwal, Ashish",
    booktitle = "Proceedings of the 61st Annual Meeting of the Association for Computational Linguistics (Volume 1: Long Papers)",
    year = "2023",
    address = "Toronto, Canada",
    publisher = "Association for Computational Linguistics",
    url = "https://aclanthology.org/2023.acl-long.557/",
    doi = "10.18653/v1/2023.acl-long.557",
    pages = "10014--10037",
}

@misc{shao2024deepseekmath,
      title={DeepSeekMath: Pushing the Limits of Mathematical Reasoning in Open Language Models}, 
      author={Zhihong Shao and Peiyi Wang and Qihao Zhu and Runxin Xu and Junxiao Song and Xiao Bi and Haowei Zhang and Mingchuan Zhang and Y. K. Li and Y. Wu and Daya Guo},
      year={2024},
      eprint={2402.03300},
      archivePrefix={arXiv},
      primaryClass={cs.CL},
      url={https://arxiv.org/abs/2402.03300}, 
}

@inproceedings{rafailov2023dpo,
author = {Rafailov, Rafael and Sharma, Archit and Mitchell, Eric and Ermon, Stefano and Manning, Christopher D. and Finn, Chelsea},
title = {Direct preference optimization: your language model is secretly a reward model},
year = {2023},
publisher = {Curran Associates Inc.},
address = {Red Hook, NY, USA},
booktitle = {Proceedings of the 37th International Conference on Neural Information Processing Systems},
articleno = {2338},
numpages = {14},
location = {New Orleans, LA, USA},
series = {NIPS '23}
}

@misc{pipitone2024legalbenchrag,
      title={LegalBench-RAG: A Benchmark for Retrieval-Augmented Generation in the Legal Domain}, 
      author={Nicholas Pipitone and Ghita Houir Alami},
      year={2024},
      eprint={2408.10343},
      archivePrefix={arXiv},
      primaryClass={cs.AI},
      url={https://arxiv.org/abs/2408.10343}, 
}

@inproceedings{jiang2025deepretrieval,
title={DeepRetrieval: Hacking Real Search Engines and Retrievers with Large Language Models via Reinforcement Learning},
author={Pengcheng Jiang and Jiacheng Lin and Lang Cao and Runchu Tian and SeongKu Kang and Zifeng Wang and Jimeng Sun and Jiawei Han},
booktitle={Second Conference on Language Modeling},
year={2025},
url={https://openreview.net/forum?id=u9JXu4L17I}
}

@article{li2026thinkthenrewrite,
author = {Li, Ang and Shi, Yufei and Si, Yuxuan and Wu, Yiquan and Cai, Ming and Tan, Xu and Wang, Yi and Sun, Changlong and Liu, Xiaozhong and Kuang, Kun},
year = {2026},
month = {03},
pages = {15045-15053},
title = {Think Then Rewrite: Reasoning Enhanced Query Rewriting for Domain Specific Retrieval},
volume = {40},
journal = {Proceedings of the AAAI Conference on Artificial Intelligence},
doi = {10.1609/aaai.v40i18.38527}
}

@inproceedings{koreeda-manning-2021-contractnli,
    title = "{C}ontract{NLI}: A Dataset for Document-level Natural Language Inference for Contracts",
    author = "Koreeda, Yuta and Manning, Christopher",
    booktitle = "Findings of the Association for Computational Linguistics: EMNLP 2021",
    year = "2021",
    address = "Punta Cana, Dominican Republic",
    publisher = "Association for Computational Linguistics",
    url = "https://aclanthology.org/2021.findings-emnlp.164",
    pages = "1907--1919"
}

@inproceedings{wang-etal-2023-maud,
  title={{MAUD}: An Expert-Annotated Legal {NLP} Dataset for Merger Agreement Understanding},
  author={Wang, Steven and Scardigli, Antoine and Tang, Leonard and Chen, Wei and Levkin, Dmitry and Chen, Anya and Ball, Spencer and Woodside, Thomas and Zhang, Oliver and Hendrycks, Dan},
  booktitle={Proceedings of the 2023 Conference on Empirical Methods in Natural Language Processing},
  pages={16369--16382},
  address={Singapore},
  publisher={Association for Computational Linguistics},
  year={2023}
}

@inproceedings{ravichander-etal-2019-question,
  title = "Question Answering for Privacy Policies: Combining Computational and Legal Perspectives",
  author = "Ravichander, Abhilasha and Black, Alan W and Wilson, Shomir and Norton, Thomas and Sadeh, Norman",
  booktitle = "Proceedings of the 2019 Conference on Empirical Methods in Natural Language Processing and the 9th International Joint Conference on Natural Language Processing (EMNLP-IJCNLP)",
  year = "2019",
  address = "Hong Kong, China",
  publisher = "Association for Computational Linguistics",
  url = "https://aclanthology.org/D19-1500",
  doi = "10.18653/v1/D19-1500",
  pages = "4947--4958"
}

@misc{fan2026legalsearchr1,
      title={Can LLMs Time Travel? Enhancing Temporal Consistency in Legal Agentic Search through Reinforcement Learning}, 
      author={Wei Fan and Yining Zhou and Mufan Zhang and Yanbing Weng and Yiran Hu and Tianshi Zheng and Baixuan Xu and Chunyang Li and Jianhui Yang and Haoran Li and Yangqiu Song},
      year={2026},
      eprint={2605.25920},
      archivePrefix={arXiv},
      primaryClass={cs.CL},
      url={https://arxiv.org/abs/2605.25920}, 
}

@misc{liu2026searchdonotguess,
      title={Search, Do not Guess: Teaching Small Language Models to Be Effective Search Agents}, 
      author={Yizhou Liu and Qi Sun and Yulin Chen and Siyue Zhang and Chen Zhao},
      year={2026},
      eprint={2604.04651},
      archivePrefix={arXiv},
      primaryClass={cs.CL},
      url={https://arxiv.org/abs/2604.04651}, 
}

@inproceedings{ahmadian2024rloo,
  title     = {Back to Basics: Revisiting {REINFORCE}-Style Optimization for Learning from Human Feedback in {LLM}s},
  author    = {Ahmadian, Arash and Cremer, Chris and Gall{\'e}, Matthias and Fadaee, Marzieh and Kreutzer, Julia and Pietquin, Olivier and {\"U}st{\"u}n, Ahmet and Hooker, Sara},
  booktitle = {Proceedings of the 62nd Annual Meeting of the Association for Computational Linguistics},
  year      = {2024},
  url       = {https://arxiv.org/abs/2402.14740}
}

@misc{eval-harness,
  title  = {Lessons from the Trenches on Reproducible Evaluation of Language Models},
  author = {Biderman, Stella and Schoelkopf, Hailey and Sutawika, Lintang and Gao, Leo and Tow, Jonathan and Abbasi, Baber and others},
  year   = {2024},
  eprint = {2405.14782},
  archivePrefix = {arXiv},
  primaryClass  = {cs.CL},
  url    = {https://arxiv.org/abs/2405.14782}
}

@inproceedings{es2024ragas,
  title     = {{RAGAS}: Automated Evaluation of Retrieval Augmented Generation},
  author    = {Es, Shahul and James, Jithin and Espinosa-Anke, Luis and Schockaert, Steven},
  booktitle = {Proceedings of the 18th Conference of the European Chapter of the Association for Computational Linguistics: System Demonstrations},
  year      = {2024},
  url       = {https://aclanthology.org/2024.eacl-demo.16/}
}

@inproceedings{gu2024minillm,
  title     = {{MiniLLM}: Knowledge Distillation of Large Language Models},
  author    = {Gu, Yuxian and Dong, Li and Wei, Furu and Huang, Minlie},
  booktitle = {The Twelfth International Conference on Learning Representations},
  year      = {2024},
  url       = {https://openreview.net/forum?id=5h0qf7IBZZ}
}

@misc{hu2024openrlhf,
  title  = {{OpenRLHF}: An Easy-to-use, Scalable and High-performance {RLHF} Framework},
  author = {Hu, Jian and Wu, Xibin and Wang, Weixun and Zhang, Dehao and Cao, Yu and others},
  year   = {2024},
  eprint = {2405.11143},
  archivePrefix = {arXiv},
  primaryClass  = {cs.CL},
  url    = {https://arxiv.org/abs/2405.11143}
}

@misc{jung2024bco,
  title  = {Binary Classifier Optimization for Large Language Model Alignment},
  author = {Jung, Seungjae and Han, Gunsoo and Nam, Daniel Wontae and On, Kyoung-Woon},
  year   = {2024},
  eprint = {2404.04656},
  archivePrefix = {arXiv},
  primaryClass  = {cs.CL},
  url    = {https://arxiv.org/abs/2404.04656}
}

@misc{langgraph2024,
  title        = {{LangGraph}: Building Stateful, Multi-Actor Applications with {LLM}s},
  author       = {{LangChain}},
  year         = {2024},
  howpublished = {\url{https://github.com/langchain-ai/langgraph}}
}

@misc{schulman2017ppo,
  title  = {Proximal Policy Optimization Algorithms},
  author = {Schulman, John and Wolski, Filip and Dhariwal, Prafulla and Radford, Alec and Klimov, Oleg},
  year   = {2017},
  eprint = {1707.06347},
  archivePrefix = {arXiv},
  primaryClass  = {cs.LG},
  url    = {https://arxiv.org/abs/1707.06347}
}

@misc{sheng2024hybridflow,
  title  = {{HybridFlow}: A Flexible and Efficient {RLHF} Framework},
  author = {Sheng, Guangming and Zhang, Chi and Ye, Zilingfeng and Wu, Xibin and Zhang, Wang and Zhang, Ru and Peng, Yanghua and Lin, Haibin and Wu, Chuan},
  year   = {2024},
  eprint = {2409.19256},
  archivePrefix = {arXiv},
  primaryClass  = {cs.LG},
  url    = {https://arxiv.org/abs/2409.19256}
}

@inproceedings{shinn2023reflexion,
  title     = {Reflexion: Language Agents with Verbal Reinforcement Learning},
  author    = {Shinn, Noah and Cassano, Federico and Berman, Edward and Gopinath, Ashwin and Narasimhan, Karthik and Yao, Shunyu},
  booktitle = {Advances in Neural Information Processing Systems},
  volume    = {36},
  year      = {2023},
  url       = {https://arxiv.org/abs/2303.11366}
}

@misc{vonwerra2020trl,
  title        = {{TRL}: Transformer Reinforcement Learning},
  author       = {von Werra, Leandro and Belkada, Younes and Tunstall, Lewis and Beeching, Edward and Thrush, Tristan and Lambert, Nathan and Huang, Shengyi},
  year         = {2020},
  howpublished = {\url{https://github.com/huggingface/trl}}
}

@inproceedings{wang2023planandsolve,
  title     = {Plan-and-Solve Prompting: Improving Zero-Shot Chain-of-Thought Reasoning by Large Language Models},
  author    = {Wang, Lei and Xu, Wanyu and Lan, Yihuai and Hu, Zhiqiang and Lan, Yunshi and Lee, Roy Ka-Wei and Lim, Ee-Peng},
  booktitle = {Proceedings of the 61st Annual Meeting of the Association for Computational Linguistics},
  year      = {2023},
  url       = {https://aclanthology.org/2023.acl-long.147/}
}

@inproceedings{yao2023tot,
  title     = {Tree of Thoughts: Deliberate Problem Solving with Large Language Models},
  author    = {Yao, Shunyu and Yu, Dian and Zhao, Jeffrey and Shafran, Izhak and Griffiths, Thomas L. and Cao, Yuan and Narasimhan, Karthik},
  booktitle = {Advances in Neural Information Processing Systems},
  volume    = {36},
  year      = {2023},
  url       = {https://arxiv.org/abs/2305.10601}
}

@misc{allal2025smollm2,
  title  = {{SmolLM2}: When Smol Goes Big -- Data-Centric Training of a Small Language Model},
  author = {Ben Allal, Loubna and Lozhkov, Anton and Bakouch, Elie and Mart{\'i}n Bl{\'a}zquez, Gabriel and Penedo, Guilherme and others},
  year   = {2025},
  eprint = {2502.02737},
  archivePrefix = {arXiv},
  primaryClass  = {cs.CL},
  url    = {https://arxiv.org/abs/2502.02737}
}

@misc{yang2025qwen3,
  title  = {{Qwen3} Technical Report},
  author = {Yang, An and others},
  year   = {2025},
  eprint = {2505.09388},
  archivePrefix = {arXiv},
  primaryClass  = {cs.CL},
  url    = {https://arxiv.org/abs/2505.09388}
}

@inproceedings{hu2022lora,
  title     = {{L}o{RA}: Low-Rank Adaptation of Large Language Models},
  author    = {Hu, Edward J. and Shen, Yelong and Wallis, Phillip and Allen-Zhu, Zeyuan and Li, Yuanzhi and Wang, Shean and Wang, Lu and Chen, Weizhu},
  booktitle = {International Conference on Learning Representations},
  year      = {2022},
  url       = {https://openreview.net/forum?id=nZeVKeeFYf9}
}

@misc{curatorkit2026,
      title={{CuratorKIT}: Data Curation and Synthetic Data Generation for {LLM} Post-Training},
      author={Bhattacharjee, Soham and Sharma, Karun and Sankarapu, Vinay Kumar and Seth, Pratinav},
      year={2026},
      eprint={2606.21631},
      archivePrefix={arXiv},
      primaryClass={cs.CL},
      url={https://arxiv.org/abs/2606.21631}
}

@misc{skyrl2025,
  title        = {{SkyRL}: A Modular Full-Stack {RL} Library for {LLM}s},
  author       = {{NovaSky-AI}},
  year         = {2025},
  howpublished = {\url{https://github.com/NovaSky-AI/SkyRL}}
}

@misc{langchain2022,
  title        = {{LangChain}: Building Applications with {LLM}s Through Composability},
  author       = {{LangChain}},
  year         = {2022},
  howpublished = {\url{https://github.com/langchain-ai/langchain}}
}

@misc{crewai2024,
  title        = {{CrewAI}: Framework for Orchestrating Role-Playing, Autonomous {AI} Agents},
  author       = {{CrewAI}},
  year         = {2024},
  howpublished = {\url{https://github.com/crewAIInc/crewAI}}
}

@misc{mhts2025,
      title={{MHTS}: Multi-Hop Tree Structure Framework for Generating Difficulty-Controllable {QA} Datasets for {RAG} Evaluation},
      author={Lee, Jeongsoo and Kwon, Daeyong and Jin, Kyohoon and Jeong, Junnyeong and Sim, Minwoo and Kim, Minwoo},
      year={2025},
      eprint={2504.08756},
      archivePrefix={arXiv},
      primaryClass={cs.IR},
      url={https://arxiv.org/abs/2504.08756}
}

@misc{mem12025,
      title={{MEM1}: Learning to Synergize Memory and Reasoning for Efficient Long-Horizon Agents},
      author={Zhou, Zijian and Qu, Ao and Wu, Zhaoxuan and Kim, Sunghwan and Prakash, Alok and Rus, Daniela and Zhao, Jinhua and Low, Bryan Kian Hsiang and Liang, Paul Pu},
      year={2025},
      eprint={2506.15841},
      archivePrefix={arXiv},
      primaryClass={cs.CL},
      url={https://arxiv.org/abs/2506.15841}
}

@misc{bgem3,
      title={{M3-Embedding}: Multi-Linguality, Multi-Functionality, Multi-Granularity Text Embeddings Through Self-Knowledge Distillation},
      author={Chen, Jianlv and Xiao, Shitao and Zhang, Peitian and Luo, Kun and Lian, Defu and Liu, Zheng},
      year={2024},
      eprint={2402.03216},
      archivePrefix={arXiv},
      primaryClass={cs.CL},
      url={https://arxiv.org/abs/2402.03216}
}

@misc{finqa,
      title={{FinQA}: A Dataset of Numerical Reasoning over Financial Data},
      author={Chen, Zhiyu and Chen, Wenhu and Smiley, Charese and Shah, Sameena and Borova, Iana and Langdon, Dylan and Moussa, Reema and Beane, Matt and Huang, Ting-Hao and Routledge, Bryan and Wang, William Yang},
      year={2021},
      eprint={2109.00122},
      archivePrefix={arXiv},
      primaryClass={cs.CL},
      url={https://arxiv.org/abs/2109.00122}
}

@inproceedings{kwon2023vllm,
  title     = {Efficient Memory Management for Large Language Model Serving with {PagedAttention}},
  author    = {Kwon, Woosuk and Li, Zhuohan and Zhuang, Siyuan and Sheng, Ying and Zheng, Lianmin and Yu, Cody Hao and Gonzalez, Joseph E. and Zhang, Hao and Stoica, Ion},
  booktitle = {Proceedings of the 29th Symposium on Operating Systems Principles},
  year      = {2023}
}

@article{robertson2009bm25,
  title   = {The Probabilistic Relevance Framework: {BM25} and Beyond},
  author  = {Robertson, Stephen and Zaragoza, Hugo},
  journal = {Foundations and Trends in Information Retrieval},
  volume  = {3},
  number  = {4},
  pages   = {333--389},
  year    = {2009}
}

@inproceedings{cormack2009rrf,
  title     = {Reciprocal Rank Fusion Outperforms {C}ondorcet and Individual Rank Learning Methods},
  author    = {Cormack, Gordon V. and Clarke, Charles L. A. and Buettcher, Stefan},
  booktitle = {Proceedings of the 32nd International ACM SIGIR Conference on Research and Development in Information Retrieval},
  pages     = {758--759},
  year      = {2009}
}

@inproceedings{lin2004rouge,
  title     = {{ROUGE}: A Package for Automatic Evaluation of Summaries},
  author    = {Lin, Chin-Yew},
  booktitle = {Text Summarization Branches Out},
  pages     = {74--81},
  year      = {2004},
  publisher = {Association for Computational Linguistics}
}

@inproceedings{papineni2002bleu,
  title     = {{BLEU}: A Method for Automatic Evaluation of Machine Translation},
  author    = {Papineni, Kishore and Roukos, Salim and Ward, Todd and Zhu, Wei-Jing},
  booktitle = {Proceedings of the 40th Annual Meeting of the Association for Computational Linguistics},
  pages     = {311--318},
  year      = {2002}
}

@inproceedings{hagberg2008networkx,
  title     = {Exploring Network Structure, Dynamics, and Function Using {NetworkX}},
  author    = {Hagberg, Aric A. and Schult, Daniel A. and Swart, Pieter J.},
  booktitle = {Proceedings of the 7th Python in Science Conference (SciPy2008)},
  pages     = {11--15},
  year      = {2008}
}

@misc{openenv2025,
  title        = {{OpenEnv}: Agentic Execution Environments},
  author       = {{Meta PyTorch} and {Hugging Face}},
  year         = {2025},
  howpublished = {\url{https://github.com/meta-pytorch/OpenEnv}}
}

@misc{unsloth,
  title        = {{Unsloth}: Finetune {LLM}s Faster with Less Memory},
  author       = {{Unsloth AI}},
  year         = {2024},
  howpublished = {\url{https://github.com/unslothai/unsloth}}
}

@misc{litellm,
  title        = {{LiteLLM}: Call All {LLM} {API}s Using the {O}pen{AI} Format},
  author       = {{BerriAI}},
  year         = {2023},
  howpublished = {\url{https://github.com/BerriAI/litellm}}
}

@misc{chroma,
  title        = {{Chroma}: The Open-Source Embedding Database},
  author       = {{Chroma}},
  year         = {2023},
  howpublished = {\url{https://github.com/chroma-core/chroma}}
}

@misc{mcp2024,
  title        = {Model Context Protocol},
  author       = {{Anthropic}},
  year         = {2024},
  howpublished = {\url{https://modelcontextprotocol.io}}
}

@misc{taubench2024,
      title={$\tau$-bench: A Benchmark for Tool-Agent-User Interaction in Real-World Domains},
      author={Yao, Shunyu and Shinn, Noah and Razavi, Pedram and Narasimhan, Karthik},
      year={2024},
      eprint={2406.12045},
      archivePrefix={arXiv},
      primaryClass={cs.AI},
      url={https://arxiv.org/abs/2406.12045}
}

@misc{toolllm2023,
      title={{ToolLLM}: Facilitating Large Language Models to Master 16000+ Real-world {API}s},
      author={Qin, Yujia and Liang, Shihao and Ye, Yining and Zhu, Kunlun and Yan, Lan and Lu, Yaxi and Lin, Yankai and Cong, Xin and Tang, Xiangru and Qian, Bill and Zhao, Sihan and Hong, Lauren and Tian, Runchu and Xie, Ruobing and Zhou, Jie and Gerstein, Mark and Li, Dahai and Liu, Zhiyuan and Sun, Maosong},
      year={2023},
      eprint={2307.16789},
      archivePrefix={arXiv},
      primaryClass={cs.AI},
      url={https://arxiv.org/abs/2307.16789}
}

@misc{agentbench2023,
      title={{AgentBench}: Evaluating {LLM}s as Agents},
      author={Liu, Xiao and Yu, Hao and Zhang, Hanchen and Xu, Yifan and Lei, Xuanyu and Lai, Hanyu and Gu, Yu and Ding, Hangliang and Men, Kaiwen and Yang, Kejuan and Zhang, Shudan and Deng, Xiang and Zeng, Aohan and Du, Zhengxiao and Zhang, Chenhui and Shen, Sheng and Zhang, Tianjun and Su, Yu and Sun, Huan and Huang, Minlie and Dong, Yuxiao and Tang, Jie},
      year={2023},
      eprint={2308.03688},
      archivePrefix={arXiv},
      primaryClass={cs.AI},
      url={https://arxiv.org/abs/2308.03688}
}

@misc{webarena2023,
      title={{WebArena}: A Realistic Web Environment for Building Autonomous Agents},
      author={Zhou, Shuyan and Xu, Frank F. and Zhu, Hao and Zhou, Xuhui and Lo, Robert and Sridhar, Abishek and Cheng, Xianyi and Ou, Tianyue and Bisk, Yonatan and Fried, Daniel and Alon, Uri and Neubig, Graham},
      year={2023},
      eprint={2307.13854},
      archivePrefix={arXiv},
      primaryClass={cs.AI},
      url={https://arxiv.org/abs/2307.13854}
}

@software{auditkit2026,
  title  = {AuditKIT: one library to evaluate any model on any dataset and any task},
  author = {Arora, Shreeyans and Avaiya, Utsav and Lyngkhoi, Zera and
            Kadiyala, Ram Mohan Rao and Gosalia, Hem and Sankarapu, Vinay Kumar and
            Seth, Pratinav},
  year   = {2026},
  version = {1.1.2},
  url    = {https://github.com/Lexsi-Labs/AuditKIT}
}

\appendix

\section{Worked examples}
\label{sec:examples}

The scenarios below show how the components compose. Each is a condensed form of a walkthrough that runs for real in the repository's notebooks.

\subsection{Self-healing a project's own trajectories}

Listing~\ref{lst:heal} runs the recovery path over a project's own episodes, entirely in Python and without a DECIDE audit log.

\begin{lstlisting}[caption={Detect failures in a project's collected trajectories (pure Python), then classify, generate corrections, and gate a retrain on a \code{RetrainingTrigger}.}, label={lst:heal}]
from agenttune.agentic import Project
from agenttune.agentic.heal_loop import SelfHealLoop
from agenttune.agentic.heal_loop import as_sync_classifier, as_sync_generator
from agenttune.decide.closed_loop.failure_classifier import FailureClassifier
from agenttune.decide.closed_loop.training_example_generator import (
    TrainingExampleGenerator)

proj = Project(strategy=strategy, harness=harness)
# ... run episodes, e.g. proj.collect_rollout(engine, tasks) ...

failures = proj.heal()   # real FailureDetector, fed by EventLog.to_audit_records()

loop = SelfHealLoop(
    classifier=as_sync_classifier(FailureClassifier(model_name="gpt-4o-mini")),
    generator=as_sync_generator(TrainingExampleGenerator(model_name="gpt-4o-mini")),
    trainer_factory=my_trainer_factory,   # omit to just preview the dataset
)
result = loop.run(failures)
print(result["n_classified"], result["n_generated"], result["trained"])
\end{lstlisting}

\subsection{A DECIDE closed-loop daemon}

The same recovery path becomes a long-lived daemon over a DECIDE audit log in Listing~\ref{lst:fullloop}.

\begin{lstlisting}[caption={\code{FullClosedLoop} watches a DECIDE audit log and drives detect $\rightarrow$ classify $\rightarrow$ retrain $\rightarrow$ gate $\rightarrow$ deploy end to end. Every model boundary is an injected callable.}, label={lst:fullloop}]
from agenttune.decide.closed_loop.full_loop import (
    FullClosedLoop, PathAConfig, GateConfig)

loop = FullClosedLoop(
    path_a=PathAConfig(audit_log_path="audit.jsonl",
                       classifier_model="gpt-4o-mini",
                       generator_model="groq/llama-3.3-70b-versatile",
                       judge_threshold=0.6, max_revisits=3),
    retrain_job=my_retrain_job,                # list[TrainingExample] -> {"path": ...}
    build_model_runner=my_build_model_runner,  # model_path -> (input -> verdict)
    old_model_runner=my_old_model_runner,
    gate_cfg=GateConfig(backend="transformers"),
)
cycles = asyncio.run(loop.run_until(stop=lambda: should_stop, poll_interval_s=5.0))
for c in cycles:
    print(c.tick.fired, c.tick.reason, c.applied)
\end{lstlisting}

\subsection{Multi-agent GRPO with a judge panel}

Listing~\ref{lst:multiagent} composes several rollout policies and several judges into the single rollout/reward pair a GRPO run needs.

\begin{lstlisting}[caption={Two rollout nodes behind a router, two judge nodes with different rubrics, compiled into a GRPO rollout/reward pair.}, label={lst:multiagent}]
from agenttune.agentic.langgraph_orchestrator import AgentTuneGraph
from agenttune.agentic.langgraph_orchestrator import make_grpo_rollout_func
from agenttune.agentic.rewards.llm_judge import LLMJudge
from agenttune.core.backend_factory import create_agentic_trainer

graph = AgentTuneGraph()
graph.add_rollout("simple",  rollout_simple)
graph.add_rollout("complex", rollout_complex)
graph.set_router(lambda p: "complex" if "convert" in p[0].lower() else "simple")
graph.add_judge("correctness", judge_correct, criteria={"accuracy": 0.7})
graph.add_judge("clarity",     judge_clarity, criteria={"clarity": 0.3})
graph.set_final_aggregation("mean")

trainer = create_agentic_trainer(
    algorithm="grpo", model="Qwen/Qwen2.5-1.5B-Instruct",
    train_dataset=my_prompt_dataset,
    rollout_func=make_grpo_rollout_func(graph.compile_rollout()),
    reward_funcs=[graph.compile_grpo_reward()],
    output_dir="./runs/router_grpo",
)
result = trainer.train()
\end{lstlisting}

\subsection{Zero-dependency synthesis run}

The six-stage synthesis pipeline of \S\ref{sec:rag} goes end to end on CPU with injected fakes in Listing~\ref{lst:synth}.

\begin{lstlisting}[caption={The six-stage synthesis pipeline on CPU with injected fakes; swapping in \code{GroqLLMClient} and \code{BGEM3Embedder} is the only change for a production run.}, label={lst:synth}]
from agenttune.rag.synthesis.build_dataset import run_pipeline
from agenttune.rag.synthesis.llm_client import FakeLLMClient
from agenttune.rag.synthesis.embeddings import FakeEmbedder

result = run_pipeline(
    docs=docs, llm=FakeLLMClient(responder=responder), embedder=FakeEmbedder(),
    out_dir="./rag_demo", backend="sqlite",
    per_hop=3, target_per_cell=3, val_fraction=0.5,
)
# writes stage0..stage3 CSVs, dataset_grpo.jsonl, metrics.json, manifest.json
\end{lstlisting}

\subsection{Training a document-search agent}

Listing~\ref{lst:ragtrain} wires a lexical index and a search tool into a GRPO run, the configuration \S\ref{sec:legal} scales up.

\begin{lstlisting}[caption={Wire a lexical index and a search tool into a GRPO run on Qwen3-0.6B \citep{yang2025qwen3}. The \code{SearchCorpusTool} instance is passed straight through; a real model calls it mid-rollout.}, label={lst:ragtrain}]
from agenttune.rag.retrieval.sqlite_fts import SQLiteFTSBackend
from agenttune.rag.tools.search_corpus import SearchCorpusTool
from agenttune.core.backend_factory import create_agentic_trainer

backend = SQLiteFTSBackend(db_path="corpus.db")
backend.index(chunks)                        # [{"chunk_id","doc_id","title","text"}...]
tool = SearchCorpusTool(backend, top_k=3)

trainer = create_agentic_trainer(
    algorithm="grpo", model="Qwen/Qwen3-0.6B",
    tools=[tool], reward_funcs=["search_grounding_reward", "answer_correctness_reward"],
    train_dataset=train_dataset, output_dir="./runs/grpo-rag",
)
trainer.train()
\end{lstlisting}

\subsection{Collecting training data from a live pipeline}

Listing~\ref{lst:collect} turns production DECIDE runs into preference- and reward-shaped training records.

\begin{lstlisting}[caption={\code{CollectRunner} turns DECIDE pipeline runs into DPO/BCO/GRPO-shaped records, read from each episode's in-memory state rather than the audit log.}, label={lst:collect}]
from agenttune.decide import GraphRunner, CollectRunner

runner = GraphRunner.from_template("bfsi/kyc_triage", config_path="config.yaml")
runner.config["episode"]["n_episodes"] = 2
n_written = await CollectRunner(runner).run(inputs=[
    "Applicant: Jane Doe, DOB 1985-03-12, income $85,000 ...",
    "Applicant: Bob Lee, DOB 1970-11-02, income $42,000 ...",
])
# bco  -> {"input","label","verdict","pipeline_id"}
# dpo  -> {"input","chosen_output","rejected_output",...}
# grpo -> {"input","output","reward",...}
\end{lstlisting}

\subsection{Authoring a tool}

Listing~\ref{lst:tool} shows the whole tool contract: a name, a description, a parameter schema, and an \code{execute} method.

\begin{lstlisting}[caption={A tool is any \code{BaseTool} subclass; only \code{execute} is required, and \code{\_parameters} feeds the model-facing schema.}, label={lst:tool}]
from agenttune.agentic.tools.base import BaseTool, ToolResult

class UppercaseTool(BaseTool):
    name = "uppercase"
    description = "Uppercase a string."

    def _parameters(self):
        return {"type": "object",
                "properties": {"text": {"type": "string"}},
                "required": ["text"]}

    def execute(self, text: str) -> ToolResult:
        return ToolResult(success=True, output=text.upper())

# tools= accepts BaseTool instances OR plain annotated callables, interchangeably.
\end{lstlisting}

\subsection{Scoring a policy without a rollout}

Listing~\ref{lst:eval} scores a policy from a static preference dataset, with no rollout and no tool loop.

\begin{lstlisting}[caption={\code{RLEvaluator} on a DPO dataset; DPO metrics are also directly callable on precomputed (chosen, rejected) score pairs.}, label={lst:eval}]
from agenttune.eval import RLEvaluator
from agenttune.eval.metrics.dpo import WinRateMetric, RewardMarginMetric

evaluator = RLEvaluator(metrics=[WinRateMetric(), RewardMarginMetric()], device="cpu")
evaluator.evaluate_rl(policy_model=policy, reference_model=reference,
                      tokenizer=tokenizer, dataset=dpo_dataset)

WinRateMetric().compute(predictions=[(0.8, 0.3), (0.2, 0.5)], references=[])
# {'win_rate': 0.5, 'win_count': 1, 'total_pairs': 2}
\end{lstlisting}

\section{\eventlog{} event kinds}
\label{app:eventkinds}

\begin{table}[!ht]
\centering
\small
\begin{tabular}{ll}
\toprule
\code{EventKind} & Meaning \\
\midrule
\code{TEXT}          & model-emitted final text \\
\code{REASONING}     & model-emitted reasoning / thought \\
\code{TOOL\_CALL}     & a tool / action invocation \\
\code{TOOL\_RESULT}   & the tool's returned output \\
\code{OBSERVATION}   & environment / state fed to the agent \\
\code{TURN\_COMPLETE} & end of one agent turn \\
\code{REWARD}        & a reward signal (\code{scope}: step\,$\vert$\,turn\,$\vert$\,episode) \\
\code{MEMORY\_OP}     & ADD / UPDATE / DELETE / NOOP on memory \\
\bottomrule
\end{tabular}
\caption{The eight \code{EventKind} values. \code{REWARD} is the only kind with a meaningful \code{scope}.}
\label{tab:eventkinds}
\end{table}

\section{Reward-function catalog}
\label{app:rewards}

\code{REWARD\_REGISTRY} holds 26 named functions, each returning \code{list[float]}, one score per completion, over completion text (needs no model). \code{combine\_rewards(names\_or\_callables, weights)} renormalizes weights to sum to one and returns a single GRPO-compatible callable whose \code{\_\_name\_\_} records the composition.

\begin{table}[!ht]
\centering
\scriptsize
\setlength{\tabcolsep}{3pt}
\begin{tabular}{lp{4.9cm}}
\toprule
Registry key & Behavior \\
\midrule
\code{correctness\_reward}          & parses \code{*yes*}/\code{*no*}; $+0.6$ matches \code{answer}, $-1.0$ wrong, $-0.5$ no match \\
\code{structure\_reward}            & call\,$\to$\,response\,$\to$\,content shape: $+0.1$/$+0.05$ well-formed, $-0.15$ call with no response \\
\code{query\_reward}                & SQL heuristic: penalizes $>3$ queries, \code{LIMIT 1}, missing \code{WHERE} \\
\code{reward\_correct\_answer}       & numeric match: $1.5$ exact ($\pm10^{-3}$), $0.5$ within $1\%$ \\
\code{reward\_tool\_used}            & $1.0$ if any message has \code{role: tool} \\
\code{reward\_concise\_answer}       & $0.3$ if the final answer is $\le 80$ words \\
\code{format\_reward}               & $+0.1$ if \code{<answer>\dots</answer>} is present \\
\code{search\_grounding\_reward}     & \code{tool\_call\_counts} tiers: $0/0.2/0.3/0.4$ \\
\code{message\_id\_citation\_reward}  & $+0.2$ if an email address or \code{message\_id} appears \\
\code{answer\_format\_reward}        & $+0.1$ if \code{<answer>} content $\ge 10$ chars \\
\code{answer\_correctness\_reward}   & \code{<answer>} vs.\ gold: $1.0$ exact, $0.5$ substring \\
\code{placeholder\_coverage\_reward} & fraction of gold numeric tokens found as substrings \\
\code{template\_structure\_reward}   & $+0.3\times$ fraction of gold \code{**headers**} reproduced \\
\code{computation\_reward}           & $0.2$/$0.3$ for 1 / $2{+}$ \code{run\_python} calls \\
\code{numerical\_match\_reward}       & first number in \code{<answer>} vs.\ gold: $1.0$ exact, $0.5$ within $5\%$ \\
\code{exploration\_reward}           & breadth tiers on \code{tool\_call\_counts}: $0/0.1/0.2/0.3$ \\
\code{tool\_chain\_reward}            & $+0.1$ per tool name in the completion, capped $0.4$ \\
\code{summary\_written\_reward}       & $+0.2$ if a new file appears in \code{sample\_dir} \\
\code{hybrid\_prm\_reward}            & $\text{tac} - w_{\text{arr}}\text{arr} - w_{\text{lcf}}\text{lcf}$ via \code{TrajectoryEvaluator}, clamped $[0,1]$ \\
\code{distilled\_judge\_reward}       & score from a \code{text-classification} pipeline over prompt$+$completion \\
\code{sql\_format\_reward}            & $+0.2$ for markdown structure \\
\code{finqa\_format\_reward}          & FinQA-specific format score \\
\code{finqa\_sql\_grounding\_reward}   & answer numbers traceable to SQL output \\
\code{finqa\_calculator\_grounding\_reward} & answer numbers traceable to calculator steps \\
\code{finqa\_placeholder\_coverage\_reward} & unit-aware, tolerance-based coverage \\
\code{finqa\_answer\_correctness\_reward}   & tolerance-based comparison across all gold numbers \\
\bottomrule
\end{tabular}
\caption{All 26 registered reward functions. The \code{finqa\_*} entries are the numerically careful versions of the plainly-named ones and are reachable under prefixed keys.}
\label{tab:rewards}
\end{table}

\section{\eventlog{} projections (source)}
\label{app:projections}

The SFT-row projection below is the one carrying the segment-aware weighting, and is reproduced because the weight assignment is the part a reader may want to check.

\begin{lstlisting}[caption={\code{as\_dataset\_rows("sft")}: one full-tier log becomes one training row with per-message weights ($1$ reasoning, $2$ tool call, $0$ environment) and a derived loss mask.}, label={lst:sftrows}]
messages, segment_weights = [], []
for e in self.events:
    if e.kind in (EventKind.REASONING, EventKind.TEXT):
        messages.append({"role": "assistant", "content": e.payload.get("text", "")})
        segment_weights.append(1.0)
    elif e.kind is EventKind.TOOL_CALL:
        for call in self._unwrap_tool_calls(e.payload.get("action") or {}):
            messages.append({"role": "assistant",
                             "content": f"<tool_call>{json.dumps(call)}</tool_call>"})
            segment_weights.append(2.0)
    elif e.kind is EventKind.TOOL_RESULT:
        messages.append({"role": "tool", "content": str(e.payload.get("output", ""))})
        segment_weights.append(0.0)
    elif e.kind is EventKind.OBSERVATION:
        messages.append({"role": "user", "content": str(e.payload.get("text", ""))})
        segment_weights.append(0.0)
return [{"messages": messages, "segment_weights": segment_weights,
        "loss_mask": [w == 0.0 for w in segment_weights]}]
\end{lstlisting}

\section{Building and checking a harness}
\label{app:harness}

\begin{lstlisting}[caption={A \code{DictToolHarness} run, the event sequence it produces, and the conformance bench that checks the harness's declared capabilities rather than trusting them.}, label={lst:harness}]
from agenttune.agentic import DictToolHarness, ReActStrategy, run_episode
from agenttune.agentic.harness import run_conformance

harness = DictToolHarness({"add": lambda a, b: a + b,
                           "multiply": lambda a, b: a * b}, max_steps=5)

log = run_episode(ReActStrategy(policy, max_steps=5), harness, "what is (4+5)*3?")
# OBSERVATION  {'text': 'what is (4+5)*3?'}
# TOOL_CALL    {'action': {'name': 'add', 'arguments': {'a': 4, 'b': 5}}}
# TOOL_RESULT  {'output': 9}
# TURN_COMPLETE {'step': 1}
# ... TOOL_CALL multiply -> 27 ... TOOL_CALL finish -> '27' ...

run_conformance(harness, "probe task")
# ConformanceReport(passed=True, drift=[])
\end{lstlisting}

\section{A DECIDE environment config (excerpt)}
\label{app:config}

Every template's \code{extends:} inherits from a \code{config.yaml} like the one below; the repository's own root config pins an open-source \code{default\_model} and the zero-API-call \code{verdict\_binary} reward.

\begin{lstlisting}[caption={\code{config.yaml} sections that matter most: model choice, reward function, collection algorithm, evaluation thresholds, audit destination.}, label={lst:configyaml}]
default_model: "Qwen/Qwen2.5-1.5B-Instruct"   # llm_call/router/judge fallback
judge_model:   "claude-opus-4-1"

run_mode: inference        # inference | collect | eval | train

reward:
  default_fn: verdict_binary   # verdict_binary | judge_score | rule_pass_rate
  normalize: true
  scale: [0.0, 1.0]

collect:
  algorithm: dpo             # dpo | bco | grpo | ppo | rloo
  min_samples: 50
  output_path: ./collected_data.jsonl

eval:
  metrics: [accuracy, latency_p50, latency_p95, cost_per_decision]
  thresholds: {accuracy: 0.80, latency_p95: 5000, cost_per_decision: 0.10}

audit:   {destination: file, path: ./audit.jsonl}
destinations:
  file:     {enabled: true, path: ./decisions.jsonl, format: jsonl}
  postgres: {enabled: false, connection_string: "...", table: decisions}
  webhook:  {enabled: false, url: "...", method: POST}
\end{lstlisting}

\section{Rollout output contract}
\label{app:rollout}

\code{create\_rollout\_fn(\dots)(prompts)} returns a dict whose keys the multi-agent auto-wiring and the GRPO patch both read back:

\begin{table}[!ht]
\centering
\scriptsize
\begin{tabular}{ll}
\toprule
Key & Contents \\
\midrule
\code{trajectories}      & \code{list[Trajectory]}, full step-by-step record incl.\ tool calls \\
\code{responses}         & \code{list[str]}, final assistant text per rollout \\
\code{rewards}           & \code{list[float]}, all zero unless \code{reward\_fn=} was passed \\
\code{prompt\_ids} / \code{completion\_ids} & token ids (the GRPO contract) \\
\code{logprobs}          & per-token logprobs \\
\code{tool\_call\_counts}  & \code{list[int]}, tool calls per rollout \\
\code{env\_mask}          & which tokens came from the environment \\
\code{conversations}     & the full message list per rollout \\
\code{retrieved\_chunk\_ids} & chunk ids surfaced by a search tool \\
\bottomrule
\end{tabular}
\caption{Keys in a rollout output dict. A bare \code{rollout\_func} must return at least \code{prompt\_ids}, \code{completion\_ids}, and \code{logprobs}; the rest are forwarded to reward functions as kwargs.}
\label{tab:rolloutkeys}
\end{table}

\section{Detailed experimental configuration}
\label{app:legalcfg}

Table~\ref{tab:legalcfg} collects the configuration behind \S\ref{sec:legal}, in one place, for reproduction.

\begin{table}[!ht]
\centering
\small
\begin{tabular}{ll}
\toprule
Parameter & Value \\
\midrule
\multicolumn{2}{l}{\emph{Data}} \\
Source corpus            & 90 CUAD training contracts \citep{hendrycks2021cuad} \\
Generated candidates     & 5{,}361 \\
After verification       & 4{,}364 \\
After answerability gate & 3{,}992 (2--5 hops) \\
Train/validation split   & gold-chunk disjoint \\
\midrule
\multicolumn{2}{l}{\emph{Search-policy training}} \\
Base model               & Qwen3.5-9B \citep{qwen3.5} \\
Algorithm                & GRPO \citep{shao2024deepseekmath} \\
Adaptation               & LoRA \citep{hu2022lora}, $r = 16$, $\alpha = 32$ \\
Learning rate            & $10^{-6}$ \\
Rewards                  & answer correctness, evidence coverage, search frugality \\
Retrieval backend        & hybrid (BM25 $+$ BGE-M3, reciprocal rank fusion) \\
Decoding at evaluation   & greedy \\
\midrule
\multicolumn{2}{l}{\emph{Distillation}} \\
Teacher                  & the trained 9B search policy \\
Students                 & Qwen2.5-0.5B, Qwen2.5-1.5B \\
Objective                & SFT on \code{full}-tier trajectories, tool outputs masked \\
\midrule
Hardware                 & 1$\times$ NVIDIA RTX PRO 6000 Blackwell Max-Q \\
\bottomrule
\end{tabular}
\caption{Configuration for the legal contract-search use case of \S\ref{sec:legal}.}
\label{tab:legalcfg}
\end{table}

\section{AuditKIT integration}
\label{app:auditkit}

AuditKIT is a separate evaluation framework within the Lexsi ecosystem for evaluating and comparing AI systems across task-specific benchmarks and evaluation criteria. It provides an evaluation layer that can be used alongside AgentTune when an external evaluation and experiment-management interface is desired. AgentTune focuses on agent construction, training, optimization, and lifecycle tooling, while AuditKIT provides complementary evaluation infrastructure. The two systems can therefore be used together without coupling the agent-training implementation to a particular evaluation framework.

In an AgentTune--AuditKIT setup, an AgentTune agent can be exposed through an evaluation interface and evaluated using AuditKIT's scenario, model, metric, and result abstractions. This separation allows the same AgentTune agent to be assessed under different evaluation configurations while keeping evaluation concerns outside the training core. 

Further information and documentation for AuditKIT are available at \cite{auditkit2026}

\section{Glossary}
\label{app:glossary}

\begin{description}[style=nextline,leftmargin=0pt,itemsep=1pt,topsep=2pt]

\item[Eventlog] The normalized two-tier trajectory schema every lifecycle stage reads and writes; projects from an RL rollout, a DECIDE state, or an eval dict, and projects into SFT rows, an evaluator dict, or audit records.
\item[Light / full tier] A \code{light} \eventlog{} is observational (content only); a \code{full} one additionally carries token spans and logprobs and is the only tier that can be trained on.
\item[Agentic spine] \code{agenttune.agentic}: the build-and-improve path, an \code{AgentStrategy} plus a \code{Harness} run through a \code{Project}.
\item[Shared RL core] \code{create\_agentic\_trainer}: the one factory that performs all weight updates, TRL-backed, dispatching GRPO/PPO/DPO/RLOO/BCO with real tool-calling rollouts.
\item[Agentic distillation] Behavior cloning: fine-tune a small student on a teacher's \code{full}-tier trajectories with a segment-aware completion-only loss, so the student learns the teacher's procedure, not its retrieved text.
\item[Segment-aware loss] Per-message loss weights ($1$ reasoning/text, $2$ tool call, $0$ environment) emitted by \code{as\_dataset\_rows("sft")}.
\item[DECIDE] The YAML decision-workflow engine; compiles a template into a LangGraph state graph with an append-only audit log.
\item[Closed loop] The self-healing system: detect $\rightarrow$ classify $\rightarrow$ generate $\rightarrow$ retrain $\rightarrow$ gate $\rightarrow$ redeploy, driven by six trigger conditions and two safety gates.
\item[Deployment gate] Blocks a redeploy unless the new model beats the old on a held-out test set reconstructed from past successful runs \emph{and} does not regress on trajectory-eval score.
\item[Conformance bench] \code{run\_conformance}: exercises a harness's declared capability flags and reports drift when behavior does not match the declaration.
\item[Harness] The Gym-style environment an agent acts in; \code{DictToolHarness} (in-process) or \code{OpenEnvHarness} (sandboxed).
\item[Rollout engine] The generation backend for the tool-calling loop: \code{transformers}, \code{vllm}, or \code{api} (LiteLLM).
\end{description}

\end{document}